%% file: main.tex
\documentclass{article} 
\usepackage{ICLR/iclr2027_conference,times}
\usepackage{booktabs}

\input{ICLR/math_commands}

\usepackage{hyperref}
\hypersetup{
  hidelinks,
  pdftitle={Difficulty Calibration Across Time for Membership Inference in Diffusion Models},
  pdfauthor={Michael Joseph and Yedid Hoshen}
}
\usepackage{url}
\usepackage{multirow}
\usepackage{wrapfig}   
\usepackage{graphicx}
\usepackage{subcaption}
\usepackage{xcolor}
\usepackage{pgfplots}
\usepackage{needspace}
\pgfplotsset{compat=1.18}
\definecolor{tscNavy}{HTML}{356D9A}
\definecolor{tscCoral}{HTML}{F45B45}
\definecolor{tscTeal}{HTML}{00A88F}
\usepackage{anyfontsize}
\usepackage{amssymb} 

\usepackage{amsthm}
\newtheorem{lemma}{Lemma}

\DeclareCaptionJustification{ablationjustified}{\fussy\pretolerance=-1\tolerance=1000\emergencystretch=0.5em\relax}

\title{Difficulty Calibration Across Time for\\
Membership Inference in Diffusion Models}

\input{macros}

\author{Michael Joseph \& Yedid Hoshen \\
\normalfont The Hebrew University of Jerusalem
}

\iclrfinalcopy
\begin{document}

\maketitle
\fancyhead{}
\renewcommand{\headrulewidth}{0pt}

\input{00_abstract}
\input{01_introduction}
\input{04_related_work}
\input{02_preliminaries}
\input{03_motivation}

\input{04_method}
\input{05_experiments}
\input{06_conclusion}

\clearpage
\input{06_statements}

\bibliography{references}
\bibliographystyle{ICLR/iclr2027_conference}

\clearpage
\input{07_appendix}

\end{document}

%% file: ICLR/math_commands.tex
\usepackage{amsmath,amsfonts,bm}

\def\eqref#1{equation~\ref{#1}}

\def\1{\bm{1}}

\DeclareMathAlphabet{\mathsfit}{\encodingdefault}{\sfdefault}{m}{sl}
\SetMathAlphabet{\mathsfit}{bold}{\encodingdefault}{\sfdefault}{bx}{n}

%% file: macros.tex
\newif\ifdraft
\draftfalse

\ifdraft
\newcommand{\michael}[1]{{\color{green}[\textbf{Michael:} #1]}}
\newcommand{\yedid}[1]{{\color{cyan}[\textbf{Yedid:} #1]}}
\newcommand{\todo}[1]{{\color{red}[\textbf{TODO:} #1]}}

\else
\newcommand{\michael}[1]{}
\newcommand{\yedid}[1]{}
\newcommand{\todo}[1]{}

\fi

%% file: 00_abstract.tex
\begin{abstract}
Modern diffusion models are trained on massive datasets that often include copyrighted or private images. Membership inference attacks (MIAs) address this by testing whether a given image was in a model's training set. The standard approach scores an image by its reconstruction loss and flags low-loss images as members. This often fails because some images are simply easier to reconstruct than others, so a simple unseen image can score a lower loss than a complex training image. Difficulty calibration separates memorization from how difficult an image is to begin with. We show how to calibrate along two axes of time: training time and diffusion time. First, across training time, we introduce Reference-Model Calibration (RMC), which compares a fine-tuned model's loss against an earlier checkpoint. This cleanly isolates newly memorized data and gives near-perfect auditing, even without text captions. Since earlier checkpoints are not always available, we next present Temporal Self-Calibration (TSC), which calibrates across diffusion time using only a single model. Our key observation is that an image's difficulty persists across noise levels, whereas memorization is localized in time. By comparing the loss at a target timestep against neighboring timesteps from the same model, TSC attenuates image difficulty using the model's own trajectory. Empirically, our methods beat the state of the art by a wide margin, and unlike most existing attacks, generalize across different model parameterizations such as  $x$-, $v$-, and $\epsilon$-prediction models.
Code is available at \url{https://github.com/MichaelJoseph777/TSC}.
\end{abstract}

%% file: 01_introduction.tex
\section{Introduction}
Diffusion Models (DMs)~\citep{song2019generative,ho2020denoising,song2020score} are currently the method of choice for high-fidelity image generation~\citep{rombach2022high, ramesh2022hierarchical}. As their success relies heavily on massive, uncurated web-scale datasets, these models often inadvertently absorb personal photographs, copyrighted artworks, and proprietary content. This indiscriminate data collection creates privacy and copyright risks, with recent studies~\citep{carlini2023extracting,somepalli2023diffusion} demonstrating that DMs can memorize and reproduce specific training examples during generation. Consequently, it has become increasingly important to reliably audit the underlying training corpora of these generative models.

Membership Inference Attacks (MIAs)~\citep{shokri2017membership} audit this risk by predicting whether a specific sample was part of the training set. While typically framed as a security vulnerability, MIAs can also enable creators to verify whether their proprietary content was used for training. These attacks exploit statistical differences in model behavior on members versus non-members~\citep{yeom2018privacy}. Although well-studied for earlier generative models such as GANs~\citep{goodfellow2020generative} and VAEs~\citep{kingma2013auto}, with several attacks developed for these settings~\citep{hayes2017logan,hilprecht2019monte,chen2020gan}, the iterative denoising process of diffusion models makes traditional MIAs considerably less effective~\citep{duan2023diffusion}, motivating approaches tailored specifically to the diffusion process.

Probably the most standard approach to MIAs is evaluating the models loss on the target examples, and predicting the low loss examples as members. Early approaches applied this to DMs by directly measuring the reconstruction error at a single, isolated timestep~\citep{matsumoto2023membership}. While followup work improved upon this ~\citep{duan2023diffusion, kong2024efficient}, a fundamental obstacle limits the accuracy of all these approaches: 
There are images that are intrinsically easy that achieve low loss regardless of membership and vice versa. This difficulty factor confounds membership making the loss an unreliable membership metric.

To overcome this confound, we propose difficulty calibration across two time axes: \emph{training time} and \emph{diffusion time}. First, across training time, we introduce \emph{Reference-Model Calibration (RMC)} to audit fine-tuned models. By comparing a sample's fine-tuned loss against the base checkpoint loss, RMC cleanly isolates newly memorized data from general reconstruction difficult. Note that this strategy is known for Large Language Models~\citep{wang2024pandora} but is under-explored in DMs. Its limitation is that prior checkpoints are frequently unavailable, and the approach is irrelevant for pretraining corpora. To provide an unconstrained alternative, we propose \emph{Temporal Self-Calibration (TSC)}, which calibrates across diffusion time using only a single model. TSC leverages the insight that image difficulty persists across nearby noise levels, whereas the membership signal is highly localized in time. By evaluating the relative loss differences between a target timestep and its temporal neighbors, TSC uses the local trajectory as an internal baseline to suppress difficulty-induced variance. This internal calibration isolates the structural deviation of membership, successfully identifying a "difficult" member even when its absolute loss strictly exceeds that of an "easy" non-member.

We evaluate our methods across multiple model architectures, datasets, and training targets. While RMC provides state-of-the-art auditing for fine-tuned Stable Diffusion models, TSC serves as a highly powerful alternative for unconstrained settings. Across DDPM and Stable Diffusion evaluation settings, TSC consistently outperforms prior baselines. On CIFAR-10~\citep{krizhevsky2009learning}, it doubles the performance of existing attacks to reach a 62.3\% TPR@1\%FPR. Furthermore, we emphasize TSC's robustness across diverse training objectives. While existing attacks are rigidly tailored to standard noise-prediction models and severely degrade on alternative parameterizations, TSC maintains highly accurate extraction capabilities. Finally, we analyze where the membership signal physically appears along the denoising trajectory, demonstrating that the effective leakage window is intrinsically tied to the underlying training objective.

Our contributions are:
\begin{itemize}
    \item We propose \emph{TSC}, a generalized method for robust membership inference. By evaluating relative loss differences across adjacent timesteps, TSC uses the local trajectory as an internal baseline to mitigate image difficulty and isolate membership signal.
    \item We introduce \emph{RMC} for auditing fine-tuned models. To the best of our knowledge, we are the first to systematically analyze and demonstrate the effectiveness of prior-checkpoint calibration of single-timestep losses for membership inference in diffusion models.
    \item We comprehensively evaluate both methods, achieving state-of-the-art performance across diverse architectures and datasets. Notably, TSC consistently yields massive improvements over prior trajectory-based baselines, frequently doubling the true positive rate at strict false-positive thresholds.
    \item We provide a detailed analysis of how the effective denoising timesteps for membership inference shift across different training objectives, extending robust membership inference beyond the standard $\epsilon$-prediction, $\epsilon$-loss paradigm.
\end{itemize}

%% file: 04_related_work.tex
\section{Related Work}
\paragraph{Membership inference attacks.}
Membership inference attacks (MIAs), originally formalized by \citet{shokri2017membership}, aim to determine whether a specific data record was included in a target model's training set. MIAs are traditionally categorized by the adversary's capabilities: white-box attacks assume access to model parameters, activations, or gradients~\citep{nasr2019comprehensive}, whereas black-box attacks operate under a more restrictive setting, relying exclusively on the model's observable outputs~\citep{choquette2021label}. Gray-box attacks occupy an intermediate threat model, granting access to quantities such as loss values or intermediate model states without requiring gradient-level information. In diffusion-model auditing, this distinction is often methodological: attacks that query intermediate denoising states generally assume local access to the model while avoiding the substantially greater computational cost of full white-box analysis. Most such query-based MIAs are \textit{threshold-based attacks}: they define a continuous membership score $\mathcal{S}_\theta(x)$, such as reconstruction loss or prediction error, for a target model $\theta$ and candidate sample $x$, and classify the sample as a member when ${\mathcal{S}_\theta(x)<\tau}$~\citep{yeom2018privacy,carlini2022membership}. The threshold $\tau$ can be empirically calibrated to optimize attack performance, often subject to a desired False Positive Rate (FPR). More sophisticated attacks additionally use shadow models to calibrate the score distribution~\citep{shokri2017membership,carlini2022membership}.

\paragraph{MIAs on diffusion models.} 
Auditing diffusion models requires analyzing the iterative denoising trajectory. While \citet{carlini2022membership} achieved high precision with the Likelihood Ratio Attack (LiRA), their approach requires training numerous shadow models, making it computationally prohibitive. Consequently, efficient threshold-based attacks have become the standard. \citet{matsumoto2023membership} introduced the Naive loss attack, which utilizes the diffusion training loss at specific timesteps as the membership score. To extract a more robust signal, \citet{duan2023diffusion} proposed SecMI, which evaluates the reconstruction error along a deterministic trajectory via DDIM inversion, effectively isolating the membership signal from the high variance of stochastic forward noise. \citet{kong2024efficient} introduced the Proximal Initialization Attack (PIA), which initializes the forward process with a known deterministic noise vector near the clean data sample. PIA infers membership by evaluating the model's noise estimation against this explicit ground-truth trajectory, achieving competitive accuracy with a reduced query budget. Finally, for conditional text-to-image models, \citet{zhai2024membership} proposed CLiD, which measures the discrepancy between conditional and unconditional diffusion losses to expose memorization specific to the text-image pair.

%% file: 02_preliminaries.tex
\section{Methodology}
\label{sec:method}

\subsection{Preliminaries}
\label{sec:prelimineries}
\paragraph{Generative trajectories and log-SNR.}
Both diffusion~\citep{ho2020denoising,song2020score} and flow-matching~\citep{lipman2022flow} models corrupt a clean sample $x$ with Gaussian noise $\epsilon \sim \mathcal{N}(0,I)$ into noisy states $z_t = \alpha_t x + \sigma_t \epsilon$ over continuous time $t$. Diffusion defines $(\alpha_t, \sigma_t)$ via a variance schedule, whereas linear flow matching introduced by ~\citet{liu2022flow} sets $\alpha_t = t$ and $\sigma_t = 1-t$ for $t \in [0,1]$, yielding $z_t = tx + (1-t)\epsilon$ with target velocity $v = x - \epsilon$. Because native time variables differ across model families, we align all trajectories using the log-signal-to-noise ratio (log-SNR):
\begin{equation}
    \lambda(t) = \log(\alpha_t^2 / \sigma_t^2)
\end{equation} 
which simplifies to $\lambda(t) = 2\log(t/(1-t))$ for linear flow, mapping noise-dominated ($\lambda \to -\infty$) to data-dominated ($\lambda \to \infty$) states.

\paragraph{Prediction and loss parameterizations.}
A denoising network $f_\theta(z_t, t, c)$ (with optional conditioning $c$) predicts a native target $p \in \{x, v, \epsilon\}$. Because $x$, $\epsilon$, and $v$ are deterministically linked via $z_t$, any native prediction $\widehat{p}_\theta$ can be algebraically converted to evaluate the squared denoising error:
\begin{equation}
    L_\theta(x, \lambda) = \| q - \widehat{q}_\theta(z_t, t, c) \|_2^2, \;\;\; q\in{\{x,v,\epsilon\}}
    \label{eq:target_loss}
\end{equation} 
all nine $(p,q)$ combinations~\citep{li2026back} are detailed in Appendix~\ref{app:pred-loss-spaces}. Evaluating $L_\theta(x, \lambda)$ across $\lambda$ for a fixed sample $x$ yields its \emph{denoising-loss trajectory}.

%% file: 03_motivation.tex
\subsection{Motivation}
\label{sec:motivation}

Most diffusion MIAs rely on a standard premise: models denoise training data more accurately than unseen data. However, absolute denoising loss varies wildly due to \emph{image difficulty} factors like high-frequency details~\citep{lian2025enhancing}, complex layouts, or rare semantic concepts that make reconstruction harder. Consequently, an "easy" non-member often incurs a lower loss than a "difficult" member, deeply entangling the two populations (Fig.~\ref{fig:motivation-hardness}). Disentangling membership from this sample difficulty requires evaluating candidates against an image-specific reference rather than relying on absolute loss.
\input{figures/motivation_hardness}

\paragraph{Calibration across training time.} For models fine-tuned from a foundational base checkpoint, the original model provides a natural reference across \emph{training time}. It establishes an accurate baseline of each image's difficulty \emph{before} the audited fine-tuning stage. As shown for Stable Diffusion on MS-COCO~\citep{lin2014microsoft} (Fig.~\ref{fig:motivation-reference}), the losses produced by the two checkpoints are highly correlated. However, subtracting the pretrained loss to compute the fine-tuning residual reveals a stark separation: members systematically shift toward lower fine-tuned loss, while non-members remain concentrated near zero.

Table~\ref{tab:reference-calibration-real} demonstrates the empirical strength of this baseline against CLiD. By isolating the loss residual and thereby neutralizing image difficulty, a minimal two-query, uncaptioned reference evaluation heavily outperforms a 15-query, fully captioned, CLiD attack. Expanding to a comparable 20-query budget yields near-perfect auditing across all datasets. These results demonstrate that separating the membership signal from baseline difficulty is crucial for strict low-FPR auditing, directly motivating our reference-free temporal approach. Appendix~\ref{app:rmc-noise-sweep} reports the noise-level sweep.

\input{figures/motivation_reference}

\begin{table}[t]
    \centering
    \caption{\textbf{RMC on realistic Stable Diffusion fine-tunes.}
    Reference calibration is evaluated without captions at $\lambda=1$ log-SNR using
    either one or 10 paired noise draws. CLiD uses the ground-truth caption. Parentheses provide the number of
    model queries per candidate.}
    \label{tab:reference-calibration-real}
    \setlength{\tabcolsep}{3.2pt}
    \renewcommand{\arraystretch}{1.12}
    \resizebox{\linewidth}{!}{%
    \begin{tabular}{@{}l@{\quad}lccc ccc ccc@{}}
        \toprule
        & & \multicolumn{3}{c}{Pok\'emon}
        & \multicolumn{3}{c}{MS COCO}
        & \multicolumn{3}{c}{Flickr} \\
        \cmidrule(lr){3-5}\cmidrule(lr){6-8}\cmidrule(lr){9-11}
        \multicolumn{2}{l}{Method}
        & AUC & ASR & TPR@1\%FPR
        & AUC & ASR & TPR@1\%FPR
        & AUC & ASR & TPR@1\%FPR \\
        \midrule
        CLiD & ($q=15$)
        & 96.98 & 91.83 & 64.75
        & 95.55 & 88.72 & 44.04
        & 95.22 & 88.16 & 47.14 \\
        \midrule
        \multirow{2}{*}{RMC} & ($q=2$)
        & \underline{99.44} & \underline{96.51} & \underline{90.63}
        & \underline{99.88} & \underline{98.52} & \underline{96.60}
        & \underline{99.80} & \underline{98.14} & \underline{96.86} \\
        & ($q=20$)
        & \textbf{99.73} & \textbf{97.84} & \textbf{96.64}
        & \textbf{99.94} & \textbf{99.04} & \textbf{98.52}
        & \textbf{99.96} & \textbf{99.50} & \textbf{99.82} \\
        \bottomrule
    \end{tabular}%
    }
\end{table}

\paragraph{Calibration across diffusion time.}
While a pretrained checkpoint provides an exceptionally strong baseline, assuming access to one is often unrealistic and irrelevant for auditing pretraining corpus. To achieve difficulty calibration without privileged external models, our primary method instead calibrates across \emph{diffusion time}, leveraging the denoising trajectory itself as a self-contained internal reference. Because image difficulty generally evolves smoothly across nearby noise levels, adjacent timesteps provide an image-specific estimate of the local loss baseline. Membership, however, often manifests as a localized deviation from this baseline. As illustrated in Figure~\ref{fig:motivation-temporal}, comparing each loss against its local temporal reference largely cancels out overall image difficulty, successfully separating members from non-members based purely on trajectory shape.

\input{figures/motivation_temporal}

%% file: figures/motivation_hardness.tex
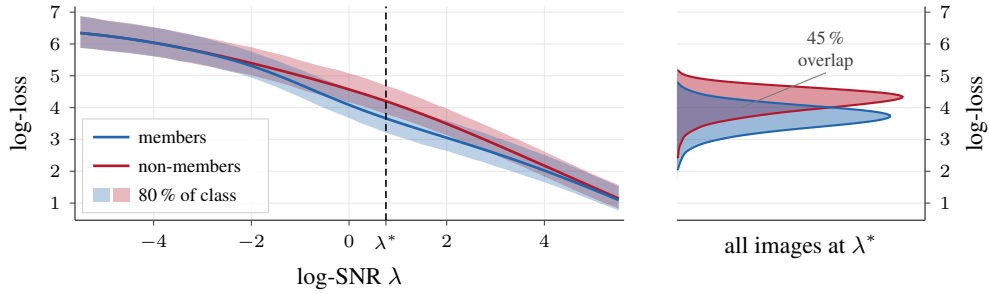
\begin{figure}[t]
  \centering
  \definecolor{mhMem}{HTML}{2166AC}
  \definecolor{mhNon}{HTML}{B2182B}
  \pgfplotsset{
    mh axis/.style={scale only axis, height=0.2025\linewidth, clip=true,
      grid=major, grid style={black!9, line width=0.25pt},
      tick label style={font=\scriptsize}, label style={font=\footnotesize},
      tick align=outside, major tick length=2pt,
      every tick/.style={black!75, line width=0.4pt},
      axis line style={black!75, line width=0.5pt}},
    mh band/.style={draw=none, fill opacity=0.3, forget plot},
    mh dens/.style={draw=none, fill opacity=0.45, forget plot},
    mh line/.style={line width=1.0pt, forget plot},
    mh edge/.style={line width=0.8pt, forget plot},
  }
  \begin{tikzpicture}
    \begin{axis}[mh axis, name=mhMain, width=0.52\linewidth,
        xmin=-5.61, xmax=5.632, ymin=0.46, ymax=7.18,
        axis lines*=left,
        xtick={-4,-2,0,2,4,0.756}, xticklabels={$-4$,$-2$,$0$,$2$,$4$,$\lambda^*$},
        ytick={1,2,3,4,5,6,7},
        xlabel={log-SNR $\lambda$}, ylabel={log-loss}]
      \addplot[mh band, fill=mhNon] coordinates {(-5.499,6.874) (-5.248,6.822) (-5.002,6.746) (-4.748,6.697) (-4.5,6.655) (-4.245,6.588) (-3.996,6.525) (-3.753,6.457) (-3.505,6.384) (-3.25,6.301) (-3.003,6.215) (-2.751,6.144) (-2.495,6.05) (-2.245,5.968) (-2.001,5.902) (-1.752,5.792) (-1.498,5.706) (-1.247,5.604) (-0.998,5.48) (-0.75,5.376) (-0.501,5.274) (-0.248,5.191) (0.001,5.077) (0.248,4.97) (0.495,4.837) (0.756,4.69) (0.997,4.573) (1.243,4.427) (1.495,4.282) (1.756,4.109) (2.008,3.97) (2.247,3.807) (2.49,3.647) (2.762,3.47) (3.013,3.3) (3.264,3.127) (3.512,2.964) (3.75,2.797) (4.019,2.605) (4.275,2.423) (4.505,2.257) (4.766,2.072) (4.987,1.914) (5.236,1.751) (5.522,1.569) (5.522,0.829) (5.236,1) (4.987,1.153) (4.766,1.305) (4.505,1.483) (4.275,1.645) (4.019,1.821) (3.75,2.008) (3.512,2.171) (3.264,2.329) (3.013,2.498) (2.762,2.662) (2.49,2.83) (2.247,2.982) (2.008,3.133) (1.756,3.262) (1.495,3.421) (1.243,3.554) (0.997,3.685) (0.756,3.793) (0.495,3.933) (0.248,4.061) (0.001,4.162) (-0.248,4.276) (-0.501,4.36) (-0.75,4.466) (-0.998,4.576) (-1.247,4.706) (-1.498,4.818) (-1.752,4.91) (-2.001,5.024) (-2.245,5.089) (-2.495,5.171) (-2.751,5.261) (-3.003,5.327) (-3.25,5.403) (-3.505,5.481) (-3.753,5.552) (-3.996,5.614) (-4.245,5.667) (-4.5,5.724) (-4.748,5.756) (-5.002,5.792) (-5.248,5.844) (-5.499,5.876)} -- cycle;
      \addplot[mh band, fill=mhMem] coordinates {(-5.499,6.87) (-5.248,6.811) (-5.002,6.74) (-4.748,6.683) (-4.5,6.626) (-4.245,6.575) (-3.996,6.515) (-3.753,6.46) (-3.505,6.361) (-3.25,6.297) (-3.003,6.215) (-2.751,6.097) (-2.495,5.988) (-2.245,5.885) (-2.001,5.763) (-1.752,5.621) (-1.498,5.488) (-1.247,5.335) (-0.998,5.193) (-0.75,5.037) (-0.501,4.884) (-0.248,4.745) (0.001,4.6) (0.248,4.439) (0.495,4.314) (0.756,4.164) (0.997,4.044) (1.243,3.933) (1.495,3.799) (1.756,3.69) (2.008,3.589) (2.247,3.462) (2.49,3.336) (2.762,3.204) (3.013,3.061) (3.264,2.934) (3.512,2.792) (3.75,2.654) (4.019,2.502) (4.275,2.343) (4.505,2.198) (4.766,2.026) (4.987,1.881) (5.236,1.716) (5.522,1.526) (5.522,0.766) (5.236,0.937) (4.987,1.086) (4.766,1.217) (4.505,1.373) (4.275,1.503) (4.019,1.645) (3.75,1.78) (3.512,1.903) (3.264,2.031) (3.013,2.145) (2.762,2.278) (2.49,2.399) (2.247,2.517) (2.008,2.638) (1.756,2.735) (1.495,2.843) (1.243,2.976) (0.997,3.09) (0.756,3.214) (0.495,3.371) (0.248,3.504) (0.001,3.675) (-0.248,3.83) (-0.501,3.981) (-0.75,4.152) (-0.998,4.327) (-1.247,4.488) (-1.498,4.659) (-1.752,4.803) (-2.001,4.947) (-2.245,5.063) (-2.495,5.154) (-2.751,5.248) (-3.003,5.353) (-3.25,5.42) (-3.505,5.472) (-3.753,5.561) (-3.996,5.608) (-4.245,5.661) (-4.5,5.707) (-4.748,5.753) (-5.002,5.795) (-5.248,5.839) (-5.499,5.874)} -- cycle;
      \addplot[mh line, mhNon] coordinates {(-5.499,6.346) (-5.248,6.306) (-5.002,6.264) (-4.748,6.215) (-4.5,6.161) (-4.245,6.102) (-3.996,6.039) (-3.753,5.974) (-3.505,5.903) (-3.25,5.827) (-3.003,5.749) (-2.751,5.665) (-2.495,5.577) (-2.245,5.488) (-2.001,5.397) (-1.752,5.303) (-1.498,5.204) (-1.247,5.104) (-0.998,5.003) (-0.75,4.9) (-0.501,4.793) (-0.248,4.683) (0.001,4.57) (0.248,4.455) (0.495,4.335) (0.756,4.203) (0.997,4.075) (1.243,3.94) (1.495,3.796) (1.756,3.642) (2.008,3.487) (2.247,3.337) (2.49,3.182) (2.762,3.004) (3.013,2.838) (3.264,2.671) (3.512,2.504) (3.75,2.343) (4.019,2.161) (4.275,1.988) (4.505,1.831) (4.766,1.654) (4.987,1.505) (5.236,1.336) (5.522,1.142)};
      \addplot[mh line, mhMem] coordinates {(-5.499,6.343) (-5.248,6.303) (-5.002,6.261) (-4.748,6.211) (-4.5,6.157) (-4.245,6.097) (-3.996,6.035) (-3.753,5.969) (-3.505,5.896) (-3.25,5.817) (-3.003,5.735) (-2.751,5.642) (-2.495,5.54) (-2.245,5.431) (-2.001,5.313) (-1.752,5.181) (-1.498,5.035) (-1.247,4.881) (-0.998,4.722) (-0.75,4.56) (-0.501,4.398) (-0.248,4.238) (0.001,4.085) (0.248,3.94) (0.495,3.802) (0.756,3.662) (0.997,3.538) (1.243,3.416) (1.495,3.293) (1.756,3.168) (2.008,3.048) (2.247,2.934) (2.49,2.817) (2.762,2.682) (3.013,2.556) (3.264,2.425) (3.512,2.293) (3.75,2.163) (4.019,2.011) (4.275,1.863) (4.505,1.726) (4.766,1.568) (4.987,1.432) (5.236,1.276) (5.522,1.094)};
      \draw[black, dashed, dash pattern=on 2.6pt off 1.6pt, line width=0.6pt]
        (0.756,0.46) -- (0.756,7.18);
    \end{axis}
    \begin{axis}[mh axis, name=mhSlice, width=0.235\linewidth,
        at={(mhMain.south east)}, anchor=south west, xshift=0.05\linewidth,
        xmin=0, xmax=1.256, ymin=0.46, ymax=7.18,
        axis y line*=right, axis x line*=bottom,
        xtick=\empty, ytick={1,2,3,4,5,6,7}, yticklabel pos=right,
        xlabel={all images at $\lambda^*$}]
      \addplot[mh dens, fill=mhNon] coordinates {(0,0.481) (0,0.481) (0,0.523) (0,0.565) (0,0.607) (0,0.649) (0,0.691) (0,0.733) (0,0.775) (0,0.817) (0,0.859) (0,0.901) (0,0.943) (0,0.985) (0,1.027) (0,1.069) (0,1.111) (0,1.153) (0,1.195) (0,1.237) (0,1.279) (0,1.321) (0,1.363) (0,1.405) (0,1.447) (0,1.489) (0,1.531) (0,1.573) (0,1.615) (0,1.657) (0,1.699) (0,1.741) (0,1.783) (0,1.825) (0,1.867) (0,1.909) (0,1.951) (0,1.993) (0,2.035) (0.001,2.077) (0.001,2.119) (0.001,2.161) (0.002,2.203) (0.002,2.245) (0.003,2.287) (0.003,2.329) (0.004,2.371) (0.005,2.413) (0.005,2.455) (0.006,2.497) (0.007,2.539) (0.008,2.581) (0.008,2.623) (0.009,2.665) (0.011,2.707) (0.012,2.749) (0.014,2.791) (0.017,2.833) (0.02,2.875) (0.023,2.917) (0.026,2.959) (0.029,3.001) (0.032,3.043) (0.037,3.085) (0.043,3.127) (0.051,3.169) (0.061,3.211) (0.072,3.253) (0.085,3.295) (0.099,3.337) (0.114,3.379) (0.131,3.421) (0.153,3.463) (0.18,3.505) (0.211,3.547) (0.246,3.589) (0.284,3.631) (0.325,3.673) (0.369,3.715) (0.416,3.757) (0.467,3.799) (0.521,3.841) (0.581,3.883) (0.647,3.925) (0.716,3.967) (0.785,4.009) (0.853,4.051) (0.916,4.093) (0.976,4.135) (1.031,4.177) (1.078,4.219) (1.115,4.261) (1.138,4.303) (1.142,4.345) (1.126,4.387) (1.09,4.429) (1.032,4.471) (0.955,4.513) (0.864,4.555) (0.764,4.597) (0.657,4.639) (0.55,4.681) (0.445,4.723) (0.35,4.765) (0.267,4.807) (0.198,4.849) (0.143,4.891) (0.1,4.933) (0.068,4.975) (0.044,5.017) (0.028,5.059) (0.017,5.101) (0.01,5.143) (0.006,5.185) (0.003,5.227) (0.002,5.269) (0.001,5.311) (0,5.353) (0,5.395) (0,5.437) (0,5.479) (0,5.521) (0,5.563) (0,5.605) (0,5.647) (0,5.689) (0,5.731) (0,5.773) (0,5.815) (0,5.857) (0,5.899) (0,5.941) (0,5.983) (0,6.025) (0,6.067) (0,6.109) (0,6.151) (0,6.193) (0,6.235) (0,6.277) (0,6.319) (0,6.361) (0,6.403) (0,6.445) (0,6.487) (0,6.529) (0,6.571) (0,6.613) (0,6.655) (0,6.697) (0,6.739) (0,6.781) (0,6.823) (0,6.865) (0,6.907) (0,6.949) (0,6.991) (0,7.033) (0,7.075) (0,7.117) (0,7.159) (0,7.159)} -- cycle;
      \addplot[mh dens, fill=mhMem] coordinates {(0,0.481) (0,0.481) (0,0.523) (0,0.565) (0,0.607) (0,0.649) (0,0.691) (0,0.733) (0,0.775) (0,0.817) (0,0.859) (0,0.901) (0,0.943) (0,0.985) (0,1.027) (0,1.069) (0,1.111) (0,1.153) (0,1.195) (0,1.237) (0,1.279) (0,1.321) (0,1.363) (0,1.405) (0,1.447) (0,1.489) (0,1.531) (0,1.573) (0,1.615) (0,1.657) (0,1.699) (0.001,1.741) (0.001,1.783) (0.001,1.825) (0.001,1.867) (0.002,1.909) (0.002,1.951) (0.003,1.993) (0.004,2.035) (0.006,2.077) (0.008,2.119) (0.01,2.161) (0.012,2.203) (0.015,2.245) (0.017,2.287) (0.02,2.329) (0.023,2.371) (0.026,2.413) (0.029,2.455) (0.033,2.497) (0.037,2.539) (0.043,2.581) (0.05,2.623) (0.06,2.665) (0.071,2.707) (0.084,2.749) (0.099,2.791) (0.115,2.833) (0.135,2.875) (0.158,2.917) (0.185,2.959) (0.216,3.001) (0.249,3.043) (0.285,3.085) (0.328,3.127) (0.376,3.169) (0.429,3.211) (0.487,3.253) (0.55,3.295) (0.618,3.337) (0.69,3.379) (0.763,3.421) (0.833,3.463) (0.898,3.505) (0.956,3.547) (1.005,3.589) (1.042,3.631) (1.067,3.673) (1.079,3.715) (1.078,3.757) (1.062,3.799) (1.032,3.841) (0.988,3.883) (0.928,3.925) (0.855,3.967) (0.772,4.009) (0.685,4.051) (0.598,4.093) (0.515,4.135) (0.438,4.177) (0.367,4.219) (0.301,4.261) (0.243,4.303) (0.194,4.345) (0.153,4.387) (0.122,4.429) (0.097,4.471) (0.076,4.513) (0.058,4.555) (0.043,4.597) (0.031,4.639) (0.021,4.681) (0.013,4.723) (0.008,4.765) (0.005,4.807) (0.003,4.849) (0.001,4.891) (0.001,4.933) (0,4.975) (0,5.017) (0,5.059) (0,5.101) (0,5.143) (0,5.185) (0,5.227) (0,5.269) (0,5.311) (0,5.353) (0,5.395) (0,5.437) (0,5.479) (0,5.521) (0,5.563) (0,5.605) (0,5.647) (0,5.689) (0,5.731) (0,5.773) (0,5.815) (0,5.857) (0,5.899) (0,5.941) (0,5.983) (0,6.025) (0,6.067) (0,6.109) (0,6.151) (0,6.193) (0,6.235) (0,6.277) (0,6.319) (0,6.361) (0,6.403) (0,6.445) (0,6.487) (0,6.529) (0,6.571) (0,6.613) (0,6.655) (0,6.697) (0,6.739) (0,6.781) (0,6.823) (0,6.865) (0,6.907) (0,6.949) (0,6.991) (0,7.033) (0,7.075) (0,7.117) (0,7.159) (0,7.159)} -- cycle;
      \addplot[mh edge, mhNon] coordinates {(0.005,2.413) (0.005,2.455) (0.006,2.497) (0.007,2.539) (0.008,2.581) (0.008,2.623) (0.009,2.665) (0.011,2.707) (0.012,2.749) (0.014,2.791) (0.017,2.833) (0.02,2.875) (0.023,2.917) (0.026,2.959) (0.029,3.001) (0.032,3.043) (0.037,3.085) (0.043,3.127) (0.051,3.169) (0.061,3.211) (0.072,3.253) (0.085,3.295) (0.099,3.337) (0.114,3.379) (0.131,3.421) (0.153,3.463) (0.18,3.505) (0.211,3.547) (0.246,3.589) (0.284,3.631) (0.325,3.673) (0.369,3.715) (0.416,3.757) (0.467,3.799) (0.521,3.841) (0.581,3.883) (0.647,3.925) (0.716,3.967) (0.785,4.009) (0.853,4.051) (0.916,4.093) (0.976,4.135) (1.031,4.177) (1.078,4.219) (1.115,4.261) (1.138,4.303) (1.142,4.345) (1.126,4.387) (1.09,4.429) (1.032,4.471) (0.955,4.513) (0.864,4.555) (0.764,4.597) (0.657,4.639) (0.55,4.681) (0.445,4.723) (0.35,4.765) (0.267,4.807) (0.198,4.849) (0.143,4.891) (0.1,4.933) (0.068,4.975) (0.044,5.017) (0.028,5.059) (0.017,5.101) (0.01,5.143) (0.006,5.185)};
      \addplot[mh edge, mhMem] coordinates {(0.004,2.035) (0.006,2.077) (0.008,2.119) (0.01,2.161) (0.012,2.203) (0.015,2.245) (0.017,2.287) (0.02,2.329) (0.023,2.371) (0.026,2.413) (0.029,2.455) (0.033,2.497) (0.037,2.539) (0.043,2.581) (0.05,2.623) (0.06,2.665) (0.071,2.707) (0.084,2.749) (0.099,2.791) (0.115,2.833) (0.135,2.875) (0.158,2.917) (0.185,2.959) (0.216,3.001) (0.249,3.043) (0.285,3.085) (0.328,3.127) (0.376,3.169) (0.429,3.211) (0.487,3.253) (0.55,3.295) (0.618,3.337) (0.69,3.379) (0.763,3.421) (0.833,3.463) (0.898,3.505) (0.956,3.547) (1.005,3.589) (1.042,3.631) (1.067,3.673) (1.079,3.715) (1.078,3.757) (1.062,3.799) (1.032,3.841) (0.988,3.883) (0.928,3.925) (0.855,3.967) (0.772,4.009) (0.685,4.051) (0.598,4.093) (0.515,4.135) (0.438,4.177) (0.367,4.219) (0.301,4.261) (0.243,4.303) (0.194,4.345) (0.153,4.387) (0.122,4.429) (0.097,4.471) (0.076,4.513) (0.058,4.555) (0.043,4.597) (0.031,4.639) (0.021,4.681) (0.013,4.723) (0.008,4.765) (0.005,4.807)};
      \draw[black!52, line width=0.4pt]
        (0.327,4.009) -- (0.754,5.095);
      \node[font=\scriptsize, align=center, text=black!75, inner sep=1pt, anchor=south]
        at (0.754,5.095) {45\,\% \\ overlap};
    \end{axis}
    \path (mhSlice.east) ++(0.95cm,0)
      node[rotate=90, anchor=south, font=\footnotesize] {log-loss};
    \begin{scope}[shift={(mhMain.south west)},
                   every node/.style={font=\scriptsize, anchor=west, inner sep=1pt}]
      \fill[white, fill opacity=0.92] (0.10,0.10) rectangle (2.56,1.34);
      \draw[black!18, line width=0.4pt] (0.10,0.10) rectangle (2.56,1.34);
      \draw[mhMem, line width=1.0pt] (0.24,1.10) -- (0.72,1.10);
      \node at (0.80,1.10) {members};
      \draw[mhNon, line width=1.0pt] (0.24,0.72) -- (0.72,0.72);
      \node at (0.80,0.72) {non-members};
      \fill[mhMem, fill opacity=0.3] (0.24,0.25) rectangle (0.46,0.43);
      \fill[mhNon, fill opacity=0.3] (0.50,0.25) rectangle (0.72,0.43);
      \node at (0.80,0.34) {80\,\% of class};
    \end{scope}
  \end{tikzpicture}
  \caption{\textbf{Image difficulty confounds absolute-loss thresholds.} \emph{Left:} CIFAR-10 DDPM log-loss trajectories for members (blue) and non-members (red) showing class means and 80\% intervals. \emph{Right:} Loss density at the optimal log-SNR ($\lambda^\star$). Even at maximal separability, the populations exhibit a massive 45\% overlap, rendering absolute-loss thresholds ineffective.}
  \label{fig:motivation-hardness}
\end{figure}

%% file: figures/motivation_temporal.tex
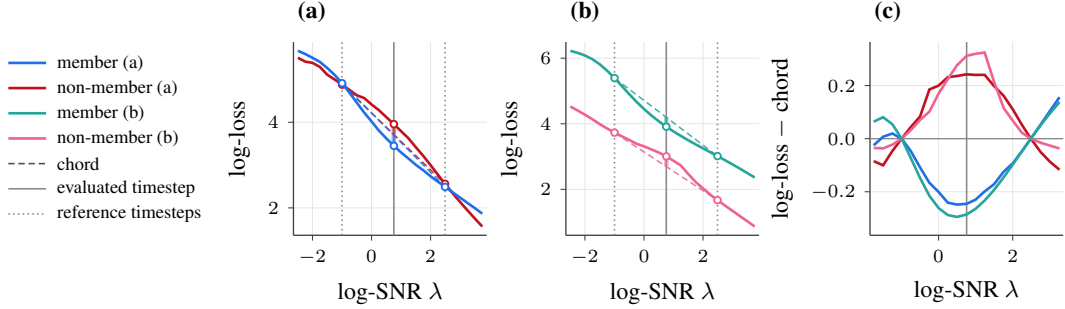
\begin{figure}[t]
  \centering
  \edef\fThreeAx{\the\dimexpr 0.183\linewidth\relax}
  \definecolor{f3MemA}{HTML}{1f6feb}\definecolor{f3NonA}{HTML}{c1121f}
  \definecolor{f3MemB}{HTML}{26a69a}\definecolor{f3NonB}{HTML}{f06292}
  \definecolor{f3Guide}{HTML}{8A8A8A}
  \pgfplotsset{
    f3 curve/.style={line width=1.0pt, forget plot},
    f3 chord/.style={line width=0.6pt, dash pattern=on 2.4pt off 1.4pt, opacity=0.85, forget plot},
    f3 bar/.style={line width=1.7pt, opacity=0.5, line cap=butt, forget plot},
    f3 pts/.style={only marks, mark=*, mark size=1.25pt, mark options={fill=white, line width=0.7pt},
                   forget plot},
  }
  \tikzset{f3 eval/.style={f3Guide, line width=0.6pt},
           f3 ref/.style={f3Guide, line width=0.6pt, dash pattern=on 0.6pt off 1.2pt}}
  \begin{tikzpicture}
    \begin{axis}[name=f3a, scale only axis, width=\fThreeAx, height=\fThreeAx, clip=true, clip mode=individual,
        axis lines*=left, axis line style={black!75, line width=0.5pt},
        grid=major, grid style={black!10, line width=0.25pt},
        tick align=outside, major tick length=2pt, every tick/.style={black!75, line width=0.4pt},
        tick label style={font=\scriptsize}, label style={font=\footnotesize},
        title style={font=\footnotesize\bfseries, at={(0,1)}, anchor=south west, inner sep=1.5pt},
        xlabel={log-SNR $\lambda$}, title={(a)},
        xmin=-2.645, xmax=3.900, ymin=1.360, ymax=5.874,
        xtick={-2,0,2,4}, ylabel={log-loss}]
      \draw[f3 eval] (axis cs:0.756,1.360) -- (axis cs:0.756,5.874);
      \draw[f3 ref] (axis cs:-0.998,1.360) -- (axis cs:-0.998,5.874);
      \draw[f3 ref] (axis cs:2.490,1.360) -- (axis cs:2.490,5.874);
      \addplot[f3 curve, f3NonA] coordinates {(-2.495,5.507) (-2.245,5.416) (-2.001,5.388) (-1.752,5.294) (-1.498,5.109) (-1.247,4.997) (-0.998,4.878) (-0.750,4.757) (-0.501,4.631) (-0.248,4.565) (0.001,4.415) (0.248,4.283) (0.495,4.123) (0.756,3.956) (0.997,3.793) (1.243,3.631) (1.495,3.421) (1.756,3.211) (2.008,2.991) (2.247,2.776) (2.490,2.562) (2.762,2.329) (3.013,2.127) (3.264,1.931) (3.512,1.742) (3.750,1.565)};
      \addplot[f3 chord, f3NonA] coordinates {(-0.998,4.878) (2.490,2.562)};
      \addplot[f3 bar, f3NonA] coordinates {(0.756,3.956) (0.756,3.720)};
      \addplot[f3 pts, f3NonA] coordinates {(-0.998,4.878) (0.756,3.956) (2.490,2.562)};
      \addplot[f3 curve, f3MemA] coordinates {(-2.495,5.669) (-2.245,5.594) (-2.001,5.506) (-1.752,5.411) (-1.498,5.266) (-1.247,5.105) (-0.998,4.912) (-0.750,4.693) (-0.501,4.459) (-0.248,4.225) (0.001,4.018) (0.248,3.806) (0.495,3.627) (0.756,3.449) (0.997,3.297) (1.243,3.157) (1.495,3.009) (1.756,2.876) (2.008,2.733) (2.247,2.612) (2.490,2.490) (2.762,2.356) (3.013,2.233) (3.264,2.109) (3.512,1.986) (3.750,1.865)};
      \addplot[f3 chord, f3MemA] coordinates {(-0.998,4.912) (2.490,2.490)};
      \addplot[f3 bar, f3MemA] coordinates {(0.756,3.449) (0.756,3.701)};
      \addplot[f3 pts, f3MemA] coordinates {(-0.998,4.912) (0.756,3.449) (2.490,2.490)};
    \end{axis}
    \begin{axis}[name=f3b, scale only axis, width=\fThreeAx, height=\fThreeAx, clip=true, clip mode=individual,
        axis lines*=left, axis line style={black!75, line width=0.5pt},
        grid=major, grid style={black!10, line width=0.25pt},
        tick align=outside, major tick length=2pt, every tick/.style={black!75, line width=0.4pt},
        tick label style={font=\scriptsize}, label style={font=\footnotesize},
        title style={font=\footnotesize\bfseries, at={(0,1)}, anchor=south west, inner sep=1.5pt},
        xlabel={log-SNR $\lambda$}, title={(b)},
        at={(f3a.south east)}, anchor=south west, xshift=0.075\linewidth,
        xmin=-2.645, xmax=3.900, ymin=0.599, ymax=6.493,
        xtick={-2,0,2,4}, ylabel={log-loss}]
      \draw[f3 eval] (axis cs:0.756,0.599) -- (axis cs:0.756,6.493);
      \draw[f3 ref] (axis cs:-0.998,0.599) -- (axis cs:-0.998,6.493);
      \draw[f3 ref] (axis cs:2.490,0.599) -- (axis cs:2.490,6.493);
      \addplot[f3 curve, f3NonB] coordinates {(-2.495,4.526) (-2.245,4.403) (-2.001,4.277) (-1.752,4.141) (-1.498,3.989) (-1.247,3.855) (-0.998,3.731) (-0.750,3.616) (-0.501,3.507) (-0.248,3.389) (0.001,3.311) (0.248,3.220) (0.495,3.125) (0.756,3.007) (0.997,2.872) (1.243,2.732) (1.495,2.473) (1.756,2.219) (2.008,2.022) (2.247,1.837) (2.490,1.671) (2.762,1.494) (3.013,1.335) (3.264,1.176) (3.512,1.019) (3.750,0.867)};
      \addplot[f3 chord, f3NonB] coordinates {(-0.998,3.731) (2.490,1.671)};
      \addplot[f3 bar, f3NonB] coordinates {(0.756,3.007) (0.756,2.701)};
      \addplot[f3 pts, f3NonB] coordinates {(-0.998,3.731) (0.756,3.007) (2.490,1.671)};
      \addplot[f3 curve, f3MemB] coordinates {(-2.495,6.225) (-2.245,6.162) (-2.001,6.082) (-1.752,5.972) (-1.498,5.815) (-1.247,5.617) (-0.998,5.394) (-0.750,5.151) (-0.501,4.906) (-0.248,4.666) (0.001,4.450) (0.248,4.256) (0.495,4.080) (0.756,3.911) (0.997,3.768) (1.243,3.632) (1.495,3.499) (1.756,3.369) (2.008,3.246) (2.247,3.131) (2.490,3.014) (2.762,2.880) (3.013,2.753) (3.264,2.623) (3.512,2.491) (3.750,2.361)};
      \addplot[f3 chord, f3MemB] coordinates {(-0.998,5.394) (2.490,3.014)};
      \addplot[f3 bar, f3MemB] coordinates {(0.756,3.911) (0.756,4.204)};
      \addplot[f3 pts, f3MemB] coordinates {(-0.998,5.394) (0.756,3.911) (2.490,3.014)};
    \end{axis}
    \begin{axis}[name=f3c, scale only axis, width=\fThreeAx, height=\fThreeAx, clip=true, clip mode=individual,
        axis lines*=left, axis line style={black!75, line width=0.5pt},
        grid=major, grid style={black!10, line width=0.25pt},
        tick align=outside, major tick length=2pt, every tick/.style={black!75, line width=0.4pt},
        tick label style={font=\scriptsize}, label style={font=\footnotesize},
        title style={font=\footnotesize\bfseries, at={(0,1)}, anchor=south west, inner sep=1.5pt},
        xlabel={log-SNR $\lambda$}, title={(c)},
        at={(f3b.south east)}, anchor=south west, xshift=0.105\linewidth,
        xmin=-1.852, xmax=3.364, ymin=-0.364, ymax=0.364,
        xtick={-2,0,2}, ytick={-0.2,0,0.2}, ylabel={log-loss $-$ chord},
        yticklabel style={/pgf/number format/fixed, /pgf/number format/precision=1,
                           /pgf/number format/fixed zerofill}]
      \draw[f3 eval] (axis cs:0.756,-0.364) -- (axis cs:0.756,0.364);
      \draw[f3Guide, line width=0.5pt] (axis cs:-1.852,0) -- (axis cs:3.364,0);
      \addplot[f3 curve, f3NonA] coordinates {(-1.752,-0.0847) (-1.498,-0.1005) (-1.247,-0.0462) (-0.998,0) (-0.750,0.0438) (-0.501,0.0833) (-0.248,0.1853) (0.001,0.1999) (0.248,0.2316) (0.495,0.2360) (0.756,0.2423) (0.997,0.2401) (1.243,0.2404) (1.495,0.1986) (1.756,0.1615) (2.008,0.1090) (2.247,0.0525) (2.490,0) (2.762,-0.0518) (3.013,-0.0879) (3.264,-0.1171)};
      \addplot[f3 curve, f3MemA] coordinates {(-1.752,-0.0250) (-1.498,0.0072) (-1.247,0.0194) (-0.998,0) (-0.750,-0.0470) (-0.501,-0.1082) (-0.248,-0.1663) (0.001,-0.2010) (0.248,-0.2411) (0.495,-0.2484) (0.756,-0.2455) (0.997,-0.2300) (1.243,-0.1988) (1.495,-0.1719) (1.756,-0.1245) (2.008,-0.0920) (2.247,-0.0475) (2.490,0) (2.762,0.0549) (3.013,0.1059) (3.264,0.1563)};
      \addplot[f3 curve, f3NonB] coordinates {(-1.752,-0.0350) (-1.498,-0.0366) (-1.247,-0.0224) (-0.998,0) (-0.750,0.0322) (-0.501,0.0702) (-0.248,0.1019) (0.001,0.1702) (0.248,0.2247) (0.495,0.2762) (0.756,0.3120) (0.997,0.3197) (1.243,0.3246) (1.495,0.2151) (1.756,0.1151) (2.008,0.0666) (2.247,0.0223) (2.490,0) (2.762,-0.0158) (3.013,-0.0264) (3.264,-0.0370)};
      \addplot[f3 curve, f3MemB] coordinates {(-1.752,0.0640) (-1.498,0.0808) (-1.247,0.0540) (-0.998,0) (-0.750,-0.0732) (-0.501,-0.1481) (-0.248,-0.2153) (0.001,-0.2616) (0.248,-0.2876) (0.495,-0.2948) (0.756,-0.2857) (0.997,-0.2644) (1.243,-0.2330) (1.495,-0.1934) (1.756,-0.1457) (2.008,-0.0962) (2.247,-0.0482) (2.490,0) (2.762,0.0519) (3.013,0.0966) (3.264,0.1378)};
    \end{axis}
    \begin{scope}[shift={(f3a.south west)}, xshift=-0.27\linewidth,
                   every node/.style={font=\scriptsize, anchor=west, align=left, inner sep=0pt}]
      \foreach \sty/\txt [count=\i] in {f3MemA/member (a), f3NonA/non-member (a),
                                        f3MemB/member (b), f3NonB/non-member (b)} {
        \draw[\sty, line width=1.0pt] (0,{2.562*(1.02-0.13*\i)}) -- ++(0.5,0);
        \node at (0.64,{2.562*(1.02-0.13*\i)}) {\txt};
      }
      \draw[black!70, line width=0.6pt, dash pattern=on 2.4pt off 1.4pt]
        (0,0.948) -- ++(0.5,0);
      \node at (0.64,0.948) {chord};
      \draw[f3 eval] (0,0.615) -- ++(0.5,0);
      \node at (0.64,0.615) {evaluated timestep};
      \draw[f3 ref] (0,0.282) -- ++(0.5,0);
      \node at (0.64,0.282) {reference timesteps};
    \end{scope}
  \end{tikzpicture}
  \caption{\textbf{Temporal self-calibration.} (a) A member/non-member pair where absolute loss correctly identifies the member.
  (b) A pair where absolute loss fails because the member is harder, yet a localized "dip" is still visible.
  (c) Subtracting the local temporal reference (the chord between the reference timesteps) removes the absolute difficulty offset.
  Both pairs collapse into a shared calibrated view, distinctly separating members from non-members by their trajectory shape.}
  \label{fig:motivation-temporal}
\end{figure}

%% file: 04_method.tex
\subsection{Difficulty Calibration Across Time}
\label{sec:membership calibration}

To mitigate the difficulty confound observed in Section~\ref{sec:motivation}, we compare a candidate's loss against an image-specific reference. We hypothesize that a significant component of difficulty acts as a persistent multiplicative factor $H(x)$. If two loss evaluations satisfy $L_a(x)=H(x)\widetilde{L}_a(x)$ and $L_b(x)=H(x)\widetilde{L}_b(x)$, where $\widetilde{L}$ represents the intrinsic loss, their ratio cleanly isolates $\widetilde{L}$ by canceling $H(x)$. While idealized, Section~\ref{sec:ablation} demonstrates this effectively reduces difficulty-induced variance in practice.

We perform this comparison in the logarithmic domain. Given a model $\theta$, image $x$, log-SNR $\lambda$, and raw loss $L_\theta(x,\lambda)$ as described in Eq.~\ref{eq:target_loss}, we define the stabilized log-loss:
\begin{equation}
    \ell_\theta(x,\lambda) = \log\!\left(L_\theta(x,\lambda)+\delta\right),
    \label{eq:log-loss}
\end{equation}
where $\delta > 0$ is a small stability constant. Subtracting two such log-losses effectively computes their log-ratio, canceling $H(x)$. We instantiate this relative calibration across two time axes: \emph{training time} (using a pretrained base checkpoint) and \emph{diffusion time} (using the image's own trajectory).

\subsubsection{Reference-Model Calibration}
\label{sec:reference-calibration}

For audited fine-tuned models ($\theta_{\text{FT}}$) with an available pretrained checkpoint ($\theta_{\text{pre}}$), the base model provides a natural estimate of an image's intrinsic difficulty prior to membership exposure. Subtracting this baseline cleanly isolates the behavioral shift caused by fine-tuning. For a fixed log-SNR $\lambda$, we define the reference-calibrated residual:
\begin{equation}
    r_{\mathrm{ref}}(x,\lambda) = \ell_{\theta_{\text{FT}}}(x,\lambda) - \ell_{\theta_{\text{pre}}}(x,\lambda).
    \label{eq:reference-calibration}
\end{equation}
As fine-tuned members incur a lower loss on $\theta_{\text{FT}}$ relative to the pretrained baseline, they systematically produce a more negative residual. We therefore use $r_{\mathrm{ref}}(x,\lambda)$ directly as our membership score.

\subsubsection{Temporal Self-Calibration}
\label{sec:temporal-calibration}
To calibrate across \emph{diffusion time}, TSC uses a candidate's own loss trajectory as an internal reference. Image difficulty tends to persist across noise levels, affecting the overall position of an image's log-loss trajectory. However, membership advantage varies along the trajectory. As illustrated in Figure~\ref{fig:temporal-window}, members tend to exhibit a greater reduction in log-loss along a certain window in the process, relative to the surrounding trajectory, than non-members.

\input{figures/method_temporal_window}

Let $\lambda^*$ denote a selected query point (e.g., the region of maximum separation), and let $\lambda^-=\lambda^*-\Delta$ and $\lambda^+=\lambda^*+\Delta$ denote two symmetric context points. We define the temporal residual as
\begin{equation}
\begin{aligned}
    r_{\mathrm{temp}}(x)
    &=
    \ell_\theta(x,\lambda^*)
    -
    \frac{1}{2}
    \left(
        \ell_\theta(x,\lambda^-)
        +
        \ell_\theta(x,\lambda^+)
    \right).
\end{aligned}
\label{eq:temporal-calibration}
\end{equation}
Geometrically, this three-point symmetric stencil measures the deviation of the log-loss at $\lambda^*$ relative to the linear chord that connects its two neighbors. 

Under the multiplicative difficulty model, the stencil weights in Eq.~\ref{eq:temporal-calibration} sum to zero, canceling the shared additive offset $\log H(x)$. Moreover, placing the context points symmetrically around $\lambda^*$ cancels linear trends and any trajectory components antisymmetric about $\lambda^*$ even at large window widths $\Delta$, eliminating residual slope variations that a single reference would retain (see Appendix~\ref{app:temporal-symmetry} and Section~\ref{sec:ablation}).

Crucially, because membership leakage is a macro-scale phenomenon, we avoid narrow windows that merely measure local curvature. Instead, we position $\lambda^*$ at the intermediate memorization peak and set a wide span $2\Delta$ so $\lambda^-$ and $\lambda^+$ fall in lower-signal regions that anchor the image's baseline trajectory. For members, memorization at $\lambda^*$ pulls the log-loss further downward relative to the linear chord connecting $\lambda^-$ and $\lambda^+$, yielding a systematically more negative residual $r_{\text{temp}}(x)$ than for non-members.

%% file: figures/method_temporal_window.tex
\begin{figure}[t]
  \centering
  \definecolor{twMem}{HTML}{1F6FEB}
  \definecolor{twNon}{HTML}{C1121F}
  \definecolor{twGuide}{HTML}{8A8A8A}
  \definecolor{twScope}{HTML}{6B6B6B}
  \pgfplotsset{
    tw axis/.style={scale only axis, height=0.25\linewidth, clip=true,
      grid=major, grid style={black!9, line width=0.25pt},
      tick label style={font=\scriptsize}, label style={font=\footnotesize},
      tick align=outside, major tick length=2pt,
      every tick/.style={black!75, line width=0.4pt}},
    tw line/.style={line width=1.0pt, forget plot},
    tw chord/.style={line width=0.7pt, dash pattern=on 2.4pt off 1.4pt, forget plot},
    tw bar/.style={line width=2.0pt, opacity=0.5, forget plot},
    tw dot/.style={only marks, mark=*, mark size=1.3pt, forget plot,
      mark options={fill=white, line width=0.7pt}},
  }
  \tikzset{
    tw eval/.style={twGuide, line width=0.5pt},
    tw ref/.style={twGuide, line width=0.5pt, dash pattern=on 0.5pt off 1.4pt},
    tw tag/.style={font=\scriptsize, text=black!60, inner sep=1.5pt, anchor=north},
    tw score/.style={font=\scriptsize, inner sep=2pt, rounded corners=1.5pt, fill=white,
      draw=#1, text=#1, line width=0.5pt},
  }
  \begin{tikzpicture}
    \begin{axis}[tw axis, name=twMain, width=0.3672\linewidth,
        xmin=-5.72, xmax=5.742, ymin=0.877, ymax=6.62,
        axis lines*=left, axis line style={black!75, line width=0.5pt},
        xtick={-4,-2,0,2,4}, ytick={1,2,3,4,5,6},
        xlabel={log-SNR $\lambda$}, ylabel={mean log-loss}]
      \draw[tw ref] (-1.752,0.877) -- (-1.752,6.62);
      \draw[tw ref] (3.264,0.877) -- (3.264,6.62);
      \draw[tw eval] (0.756,0.877) -- (0.756,6.62);
      \node[tw tag] at (-1.752,6.62) {$\lambda^-$};
      \node[tw tag] at (0.756,6.62) {$\lambda^*$};
      \node[tw tag] at (3.264,6.62) {$\lambda^+$};
      \addplot[tw line, twNon] coordinates {(-5.499,6.356) (-5.248,6.32) (-5.002,6.28) (-4.748,6.231) (-4.5,6.176) (-4.245,6.112) (-3.996,6.044) (-3.753,5.972) (-3.505,5.894) (-3.25,5.81) (-3.003,5.726) (-2.751,5.638) (-2.495,5.545) (-2.245,5.454) (-2.001,5.363) (-1.752,5.269) (-1.498,5.171) (-1.247,5.073) (-0.998,4.975) (-0.75,4.874) (-0.501,4.772) (-0.248,4.664) (0.001,4.556) (0.248,4.444) (0.495,4.328) (0.756,4.2) (0.997,4.076) (1.243,3.946) (1.495,3.806) (1.756,3.655) (2.008,3.506) (2.247,3.359) (2.49,3.208) (2.762,3.035) (3.013,2.872) (3.264,2.707) (3.512,2.543) (3.75,2.384) (4.019,2.203) (4.275,2.029) (4.505,1.872) (4.766,1.694) (4.987,1.543) (5.236,1.371) (5.522,1.174)};
      \addplot[tw chord, twNon] coordinates {(-1.752,5.269) (3.264,2.707)};
      \addplot[tw bar, twNon] coordinates {(0.756,4.2) (0.756,3.988)};
      \addplot[tw dot, draw=twNon] coordinates {(-1.752,5.269) (0.756,4.2) (3.264,2.707)};
      \addplot[tw line, twMem] coordinates {(-5.499,6.359) (-5.248,6.323) (-5.002,6.282) (-4.748,6.232) (-4.5,6.177) (-4.245,6.113) (-3.996,6.044) (-3.753,5.971) (-3.505,5.892) (-3.25,5.806) (-3.003,5.717) (-2.751,5.623) (-2.495,5.52) (-2.245,5.414) (-2.001,5.302) (-1.752,5.18) (-1.498,5.047) (-1.247,4.907) (-0.998,4.762) (-0.75,4.613) (-0.501,4.463) (-0.248,4.312) (0.001,4.168) (0.248,4.028) (0.495,3.895) (0.756,3.758) (0.997,3.636) (1.243,3.515) (1.495,3.394) (1.756,3.269) (2.008,3.149) (2.247,3.034) (2.49,2.915) (2.762,2.779) (3.013,2.649) (3.264,2.516) (3.512,2.38) (3.75,2.245) (4.019,2.088) (4.275,1.935) (4.505,1.793) (4.766,1.629) (4.987,1.488) (5.236,1.326) (5.522,1.138)};
      \addplot[tw chord, twMem] coordinates {(-1.752,5.18) (3.264,2.516)};
      \addplot[tw bar, twMem] coordinates {(0.756,3.758) (0.756,3.848)};
      \addplot[tw dot, draw=twMem] coordinates {(-1.752,5.18) (0.756,3.758) (3.264,2.516)};
      \draw[twScope, line width=0.6pt] (-0.804,2.721) rectangle (2.316,5.187);
      \coordinate (twRT) at (2.316,5.187);
      \coordinate (twRB) at (2.316,2.721);
    \end{axis}
    \begin{axis}[tw axis, name=twZoom, width=0.2328\linewidth,
        at={(twMain.south east)}, anchor=south west, xshift=0.055\linewidth,
        xmin=-0.804, xmax=2.316, ymin=2.721, ymax=5.187,
        axis line style={twScope, line width=0.6pt},
        xtick={-0.5,0,0.5,1,1.5,2}, ytick={2.75,3,3.25,3.5,3.75,4,4.25,4.5,4.75,5}, yticklabel pos=right,
        xlabel={log-SNR $\lambda$},
        yticklabel style={/pgf/number format/fixed, /pgf/number format/precision=2,
                           /pgf/number format/fixed zerofill}]
      \draw[tw eval] (0.756,2.721) -- (0.756,5.187);
      \node[tw tag, anchor=south, xshift=3pt] at (0.756,2.721) {$\lambda^*$};
      \addplot[tw line, twNon] coordinates {(-0.998,4.975) (-0.75,4.874) (-0.501,4.772) (-0.248,4.664) (0.001,4.556) (0.248,4.444) (0.495,4.328) (0.756,4.2) (0.997,4.076) (1.243,3.946) (1.495,3.806) (1.756,3.655) (2.008,3.506) (2.247,3.359) (2.49,3.208)};
      \addplot[tw chord, twNon] coordinates {(-0.998,4.884) (-0.75,4.757) (-0.501,4.63) (-0.248,4.501) (0.001,4.374) (0.248,4.248) (0.495,4.121) (0.756,3.988) (0.997,3.865) (1.243,3.739) (1.495,3.611) (1.756,3.477) (2.008,3.349) (2.247,3.227) (2.49,3.103)};
      \addplot[tw bar, twNon] coordinates {(0.756,4.2) (0.756,3.988)};
      \addplot[tw dot, draw=twNon] coordinates {(0.756,4.2)};
      \addplot[tw line, twMem] coordinates {(-0.998,4.762) (-0.75,4.613) (-0.501,4.463) (-0.248,4.312) (0.001,4.168) (0.248,4.028) (0.495,3.895) (0.756,3.758) (0.997,3.636) (1.243,3.515) (1.495,3.394) (1.756,3.269) (2.008,3.149) (2.247,3.034) (2.49,2.915)};
      \addplot[tw chord, twMem] coordinates {(-0.998,4.78) (-0.75,4.648) (-0.501,4.516) (-0.248,4.381) (0.001,4.249) (0.248,4.118) (0.495,3.987) (0.756,3.848) (0.997,3.72) (1.243,3.589) (1.495,3.455) (1.756,3.317) (2.008,3.183) (2.247,3.056) (2.49,2.927)};
      \addplot[tw bar, twMem] coordinates {(0.756,3.758) (0.756,3.848)};
      \addplot[tw dot, draw=twMem] coordinates {(0.756,3.758)};
      \node[tw score=twNon] (twLabNon) at (1.13,4.941)
        {TSC score $+0.212$};
      \node[tw score=twMem] (twLabMem) at (0.382,3.091)
        {TSC score $-0.090$};
      \draw[->, twNon, line width=0.5pt, shorten >=1pt]
        (twLabNon) -- (0.756,4.094);
      \draw[->, twMem, line width=0.5pt, shorten >=1pt]
        (twLabMem) -- (0.756,3.803);
    \end{axis}
    \draw[twScope, line width=0.45pt] (twRT) -- (twZoom.north west);
    \draw[twScope, line width=0.45pt] (twRB) -- (twZoom.south west);
    \begin{scope}[shift={(twMain.outer north west)}, xshift=-0.205\linewidth,
                   every node/.style={font=\scriptsize, anchor=west, align=left,
                                      inner sep=0pt}]
      \draw[twMem, line width=1.0pt] (0,-0.55) -- (0.52,-0.55);
      \node at (0.64,-0.55) {members\\(mean)};
      \draw[twNon, line width=1.0pt] (0,-1.30) -- (0.52,-1.30);
      \node at (0.64,-1.30) {non-members\\(mean)};
      \draw[twMem, line width=0.7pt, dash pattern=on 2.4pt off 1.4pt] (0,-1.95) -- (0.25,-1.95);
      \draw[twNon, line width=0.7pt, dash pattern=on 2.4pt off 1.4pt]
        (0.27,-1.95) -- (0.52,-1.95);
      \node at (0.64,-1.95) {chord};
      \draw[twMem, line width=2.0pt, opacity=0.5] (0,-2.50) -- (0.25,-2.50);
      \draw[twNon, line width=2.0pt, opacity=0.5] (0.27,-2.50) -- (0.52,-2.50);
      \node at (0.64,-2.50) {TSC score};
      \draw[tw eval] (0,-3.15) -- (0.52,-3.15);
      \node at (0.64,-3.15) {evaluated\\timestep};
      \draw[tw ref] (0,-3.90) -- (0.52,-3.90);
      \node at (0.64,-3.90) {reference\\timesteps};
    \end{scope}
  \end{tikzpicture}
  \caption{\textbf{TSC on STL-10-U.} Mean member and non-member log-loss trajectories (left) and a close-up of the selected window (right). The chords connect the two context points for each population, and the vertical bars at $\lambda^*$ show the residual measured by the temporal score.}
  \label{fig:temporal-window}
\end{figure}
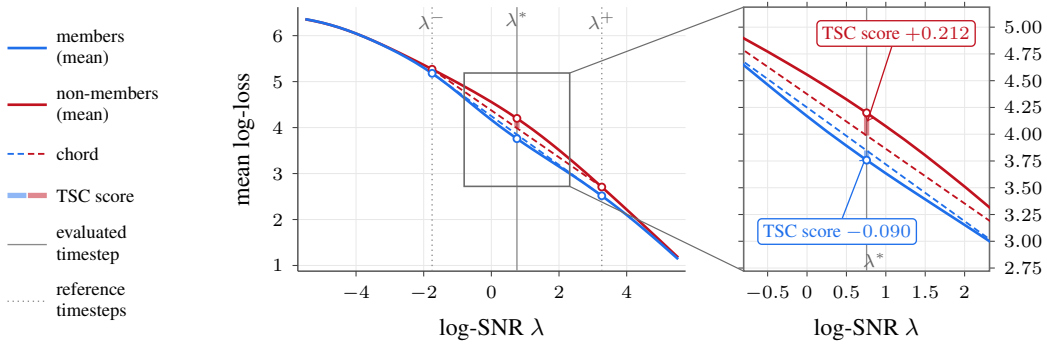

%% file: 05_experiments.tex
\section{Experiments}
\label{sec:experiments}

\subsection{Experiment Setup}
\label{sec:experiment-setup}

\paragraph{Models and datasets.}
For standard image synthesis, we evaluate DDPMs~\citep{ho2020denoising} on CIFAR-10, CIFAR-100~\citep{krizhevsky2009learning}, STL-10-U~\citep{coates2011analysis}, and Tiny ImageNet~\citep{deng2009imagenet}, as well as DDPM++ backbone~\citep{song2020score} on CIFAR-100. For text-to-image generation, we evaluate Stable Diffusion v1.4~\citep{rombach2022high} fine-tuned on the Pok\'emon~\citep{pinkney2022pokemon}, MS-COCO~\citep{lin2014microsoft}, and Flickr30k~\citep{young2014image} datasets. Complete model configurations and dataset splits are detailed in Appendix~\ref{app:experimental-details}.

\paragraph{Baselines and evaluation metrics.}
We compare TSC against established MIAs for DMs: Naive~\citep{matsumoto2023membership}, SecMI~\citep{duan2023diffusion}, PIA, and PIAN~\citep{kong2024efficient}. We measure performance across all methods using Attack Success Rate (ASR), AUC, and True Positive Rate at 1\% False Positive Rate (TPR@1\%FPR).

\paragraph{Implementation details.}
Hyperparameters for both TSC (the optimal center $\lambda^\star$) and timestep-dependent baselines (Naive, PIA) are tuned by maximizing AUC on a disjoint shadow data split. As analyzed in our results, these optimal coordinates naturally shift depending on the model's loss weighting. For TSC, we fix the temporal window at $2\Delta = 5$ log-SNR. $\lambda^*$ is fixed at $0.75$ for DDPM benchmarks and $0.25$ for SD trained models. To balance efficiency, we allocate a proportionate query budget that ensures stable evaluation. However, as shown in Fig.\ref{fig:ablation-score-query}, the attack performance can improve further with higher query budget.

\subsection{Temporal Self-Calibration Evaluations and Analysis}
\label{sec:results-temporal}

\begin{table}[t]
    \centering
    \caption{\textbf{Membership inference on DDPM benchmarks.}
    Results reported as percentages. Parentheses give the number of model queries per
    candidate. SecMI follows its official protocol.}
    \label{tab:ddpm-unconditional-results}
    \setlength{\tabcolsep}{3.2pt}
    \renewcommand{\arraystretch}{1.12}
    \resizebox{\linewidth}{!}{%
    \begin{tabular}{@{}l@{\quad}lccc ccc ccc@{}}
        \toprule
        & & \multicolumn{3}{c}{CIFAR-10}
        & \multicolumn{3}{c}{Tiny ImageNet}
        & \multicolumn{3}{c}{STL-10-U} \\
        \cmidrule(lr){3-5}\cmidrule(lr){6-8}\cmidrule(lr){9-11}
        \multicolumn{2}{l}{Method}
        & AUC & ASR & TPR@1\%FPR
        & AUC & ASR & TPR@1\%FPR
        & AUC & ASR & TPR@1\%FPR \\
        \midrule
        Naive & ($q=20$)
        & 85.39 & 78.21 & 7.35
        & 64.37 & 60.52 & 2.93
        & 64.15 & 60.50 & 2.51 \\
        SecMI & ($q=12$)
        & 88.24 & 80.91 & 10.20
        & 66.37 & 61.88 & 2.91
        & 68.12 & 63.33 & 3.03 \\
        PIA & ($q=2$)
        & \underline{91.90} & \underline{84.88} & 25.80
        & \underline{68.99} & \underline{64.25} & 4.43
        & \underline{69.46} & \underline{64.51} & 4.08 \\
        PIAN & ($q=2$)
        & 89.26 & 83.32 & \underline{30.90}
        & 67.04 & 63.44 & \underline{6.10}
        & 66.72 & 63.36 & \underline{4.32} \\
        \midrule
        TSC & ($q=15$)
        & \textbf{95.91} & \textbf{90.15} & \textbf{62.31}
        & \textbf{78.63} & \textbf{72.18} & \textbf{12.27}
        & \textbf{77.84} & \textbf{71.29} & \textbf{9.57} \\
        \bottomrule
    \end{tabular}%
    }
\end{table}

\begin{table}[t]
    \centering
    \caption{\textbf{Membership inference on Stable Diffusion fine-tunes (uncaptioned).}
    SecMI denotes its LDM variant. PIA, is evaluated both over the original paper's default timestep ($t=500$) and an average optimal timestep over shadow models.}
    \label{tab:sd-finetuned-results}
    \setlength{\tabcolsep}{3.2pt}
    \renewcommand{\arraystretch}{1.12}
    \resizebox{\linewidth}{!}{%
    \begin{tabular}{@{}l@{\quad}lccc ccc ccc@{}}
        \toprule
        & & \multicolumn{3}{c}{Pok\'emon}
        & \multicolumn{3}{c}{MS COCO}
        & \multicolumn{3}{c}{Flickr} \\
        \cmidrule(lr){3-5}\cmidrule(lr){6-8}\cmidrule(lr){9-11}
        \multicolumn{2}{l}{Method}
        & AUC & ASR & TPR@1\%FPR
        & AUC & ASR & TPR@1\%FPR
        & AUC & ASR & TPR@1\%FPR \\
        \midrule
        Naive & ($q=10$)
        & 84.84 & 77.43 & 28.37
        & 76.61 & 69.92 & 7.16
        & 77.96 & 71.66 & 10.08 \\
        SecMI & ($q=12$)
        & 83.59 & 76.47 & 15.14
        & \underline{78.33} & \underline{71.46} & 6.84
        & \underline{81.14} & \underline{74.24} & 7.40 \\
        PIA ($t=500$) & ($q=2)$ 
        & 69.86 & 66.13 & 19.71
        & 54.08 & 53.96 & 2.76
        & 54.73 & 53.24 & 1.96 \\
        PIA (optimal $t$) & ($q=2$)
        & \textbf{94.50} & \textbf{87.88} & \textbf{61.30}
        & 75.61 & 70.90 & \underline{17.36}
        & 74.18 & 69.58 & \underline{17.12} \\
        PIAN & ($q=2$)
        & 53.34 & 54.15 & 1.92
        & 69.20 & 64.08 & 13.88
        & 69.15 & 65.10 & 12.40 \\
        \midrule
        TSC & ($q=15$)
        & \underline{91.37} & \underline{84.03} & \underline{55.77}
        & \textbf{85.55} & \textbf{78.42} & \textbf{29.16}
        & \textbf{84.60} & \textbf{77.98} & \textbf{30.28} \\
        \bottomrule
    \end{tabular}%
    }
\end{table}

Table~\ref{tab:ddpm-unconditional-results} presents the performance of our method TSC against the baselines on standard image synthesis models. TSC consistently achieves state-of-the-art performance across all datasets. While it provides solid improvements in overall AUC and ASR, its most significant advantage emerges in the high-confidence regime. On CIFAR-10, TSC achieves a TPR@1\%FPR of 62.31\%, doubling the performance of the strongest baseline (PIAN at 30.90\%). This trend persists across the more complex Tiny ImageNet and STL-10-U datasets, demonstrating that our geometric measurement of the loss trajectory effectively isolates members while bounding false positives.

For text-to-image models (Table~\ref{tab:sd-finetuned-results}), we evaluate Stable Diffusion under a realistic uncaptioned threat model. Because training prompts are typically unknown to an auditor, omitting ground-truth captions prevents artificially inflating attack success via prompt-specific overfitting, directly isolating the inherent image-level membership signal.

Under this setting, TSC outperforms all baselines across every metric on MS-COCO and Flickr30k. The exception is the Pok\'emon fine-tune, where PIA prevails. We attribute this to the dataset's limited scale and visual uniformity. Such low diversity naturally minimizes the variance in image difficulty, allowing PIA's proximal measurement to perform exceptionally well without requiring relative calibration. Furthermore, our shadow-model tuning shifts PIA's optimal timestep from its official default of $t=500$~\citep{kong2024efficient} down to $t \approx 250\text{-}300$. By evaluating PIA at these tuned coordinates, our reported baselines are notably stronger than standard static evaluations.

\subsection{Impact of Prediction Parameterization and Training Target}
\label{sec:loss_temporal_localization}
Membership leakage varies along the trajectory, making timestep selection crucial.
To understand how the training objective influences the informative region,
we express the losses in a common velocity coordinate.

Under linear flow,
$x=z_t+(1-t)v$ and $\epsilon=z_t-tv$. For a fixed native head
$p\in\{x,v,\epsilon\}$ and $t\in(0,1)$, a converted velocity error
$\delta_v=\widehat v_\theta^{(p)}-v$ induces the errors
\[
    \delta_x=(1-t)\delta_v,\qquad \delta_\epsilon=-t\delta_v.
\]
Squaring these residuals gives
$\mathcal{L}_{p,s}(t)=w_s(t)\|\delta_v\|_2^2$, where
$(w_x,w_v,w_\epsilon)=((1-t)^2,1,t^2)$ depends only on the
training-loss target $s\in\{x,v,\epsilon\}$.

\begin{lemma}[Temporal Weight Ordering]
\normalfont
For a common non-degenerate timestep distribution $\rho$ on $(0,1)$,
the explicit-weight barycenters
$c_s\triangleq\mathbb{E}_\rho[T w_s(T)]/\mathbb{E}_\rho[w_s(T)]$
satisfy $c_x<c_v<c_\epsilon$. Under uniform sampling,
$(c_x,c_v,c_\epsilon)=(1/4,1/2,3/4)$.
Proof in Appendix~\ref{app:barycenter_proof}.
\end{lemma}

In this coordinate, $x$-loss emphasizes noisy timesteps and $\epsilon$-loss
emphasizes cleaner timesteps, while $v$-loss adds no temporal weighting beyond
$\rho$. This ordering concerns explicit weights, it does not determine leakage
peaks, which also depends on learned errors and optimization dynamics.

We examine whether the measured leakage follows the same ordering across nine
separately trained CIFAR-100 models spanning all $(p,s)$ combinations.
Single-timestep AUC sweeps (Fig.~\ref{fig:edm_loss_localization},
Appendix~\ref{app:loss_auc_curves}) show that peak leakage shifts toward higher
log-SNR as the training target changes from $x$ to $v$ to $\epsilon$, whereas
changing the native head $p$ has little effect on the peak location.

\begin{table}[t]
    \centering
    \caption{\textbf{Robustness across prediction and loss parameterizations.}
    TPR@1\%FPR (\%) for unaugmented CIFAR-100 DDPM++ FM models. 
    The selected-$\lambda^\star$ tuple lists the shared operating log-SNR under 
    $x$-, $v$-, and $\epsilon$-losses. SecMI follows its default protocol.}
    \label{tab:pred-loss-results}
    \setlength{\tabcolsep}{2.6pt}
    \renewcommand{\arraystretch}{1.12}
    \resizebox{\linewidth}{!}{%
    \begin{tabular}{@{}l@{\quad}l c ccc ccc ccc@{}}
        \toprule
        & & \multirow{2}{*}{\shortstack{Selected $\lambda^\star$\\
            ($x/v/\epsilon$ loss)}}
        & \multicolumn{3}{c}{$x$-loss}
        & \multicolumn{3}{c}{$v$-loss}
        & \multicolumn{3}{c}{$\epsilon$-loss} \\
        \cmidrule(lr){4-6}\cmidrule(lr){7-9}\cmidrule(lr){10-12}
        \multicolumn{2}{l}{Method} &
        & $x$-pred & $v$-pred & $\epsilon$-pred
        & $x$-pred & $v$-pred & $\epsilon$-pred
        & $x$-pred & $v$-pred & $\epsilon$-pred \\
        \midrule
        Naive & ($q=20$) & $-0.75/0.25/1.50$
        & \underline{6.48} & \underline{6.05} & \underline{6.74}
        & \underline{34.54} & 34.28 & 34.56
        & 27.52 & 29.78 & 28.29 \\
        PIA & ($q=2$) & $-0.75/0.25/1.75$
        & 2.27 & 4.07 & 6.22
        & 1.01 & 52.48 & 54.84
        & 31.69 & 74.61 & 75.85 \\
        PIAN & ($q=2$) & $-0.50/0.50/1.75$
        & 2.41 & 6.00 & 5.69
        & 1.06 & \underline{56.68} & \underline{58.36}
        & \underline{72.11} & \underline{77.77} & \underline{76.73} \\
        SecMI & ($q=12$) & default
        & 0.94 & 1.10 & 1.19
        & 5.55 & 5.57 & 5.26
        & 29.25 & 35.40 & 30.82 \\
        \midrule
        TSC & ($q=15$) & $-0.50/0.50/1.75$
        & \textbf{30.28} & \textbf{28.74} & \textbf{31.26}
        & \textbf{81.31} & \textbf{80.56} & \textbf{80.68}
        & \textbf{91.22} & \textbf{93.22} & \textbf{92.75} \\
        \bottomrule
    \end{tabular}%
    }
\end{table}

Table~\ref{tab:pred-loss-results} evaluates attacks across all nine models.
Consistent with the observed ordering, TSC's center increases from
$\lambda^\star=-0.50$ ($x$-loss) to $0.50$ ($v$-loss) and $1.75$
($\epsilon$-loss). TSC achieves the highest TPR@1\%FPR in every configuration,
improving over the strongest baseline by $23.84$ percentage points on average.
Under $x$-loss, baselines remain below $7\%$ TPR while TSC reaches $29$--$31\%$.
These results support adapting the operating region to the training objective
and demonstrate TSC's effectiveness across the tested parameterizations.

\subsection{Ablations}
\label{sec:ablation}

\par\smallskip
\noindent\begin{minipage}[c]{0.52\linewidth}
    \noindent\phantomsection\label{abl:score-query}%
    \textbf{Score domain and query budget.} We ablate the loss domain and query budget $q$ (Fig.~\ref{fig:ablation-score-query}). Evaluating TSC on logarithmic rather than raw loss reveals a stark contrast: log-loss TSC reaches nearly 26\% TPR@1\%FPR, while raw-loss TSC falls below the Naive baseline. This strengthens our hypothesis that sample difficulty acts as a multiplicative confounder. The log transformation converts this into an additive constant, allowing TSC to effectively cancel it out.
    
    \vspace{1ex}
    Furthermore, as queries increase from $q=3$ to $q=45$, log-loss TSC scales efficiently. Additional noise draws at the same three query levels improve estimation of the temporal contrast. In this experiment, the Naive attack remains near 4\%, gaining little from the additional loss evaluations.
\end{minipage}\hfill
\begin{minipage}[c]{0.45\linewidth}
    \captionsetup{type=figure}
    \centering
    \begin{tikzpicture}
        \begin{axis}[
            width=\linewidth,
            height=0.75\linewidth,
            xmode=log,
            log basis x=10,
            xmin=2.7,
            xmax=52,
            ymin=0,
            ymax=0.38, 
            xtick={3,6,15,30,45},
            xticklabels={3,6,15,30,45},
            ytick={0,0.05,0.10,0.15,0.20,0.25,0.30,0.35},
            yticklabels={0.00,0.05,0.10,0.15,0.20,0.25,0.30,0.35},
            xlabel={Queries},
            ylabel={TPR@1\%FPR},
            axis x line*=bottom,
            axis y line*=left,
            tick align=outside,
            major grid style={draw=black!12, line width=0.35pt},
            grid=major,
            minor tick num=0,
            tick label style={font=\scriptsize},
            label style={font=\small},
            legend style={
                at={(0.02,0.98)}, 
                anchor=north west,
                draw=none,
                fill=none,
                font=\tiny, 
                row sep=0pt, 
                inner sep=1pt, 
            },
            legend cell align=left,
        ]
            \addplot[
                color=tscTeal,
                line width=1.3pt,
                mark=*,
                mark size=2.2pt,
            ] coordinates {
                (3,0.1494575) (6,0.1910250) (15,0.2327525)
                (30,0.2556125) (45,0.2598150)
            };
            \addlegendentry{\textcolor{tscTeal}{TSC (log loss)}}

            \addplot[
                color=tscNavy,
                line width=1.3pt,
                dashed,
                mark=*,
                mark options={solid},
                mark size=2.2pt,
            ] coordinates {
                (3,0.0426025) (6,0.0416125) (15,0.0420450)
                (30,0.0419550) (45,0.0419800)
            };
            \addlegendentry{\textcolor{tscNavy}{Naive (loss)}}

            \addplot[
                color=tscCoral,
                line width=1.3pt,
                densely dotted,
                mark=square*,
                mark options={solid},
                mark size=2.2pt,
            ] coordinates {
                (3,0.0228550) (6,0.0230475) (15,0.0247225)
                (30,0.0251175) (45,0.0255250)
            };
            \addlegendentry{\textcolor{tscCoral}{TSC (raw loss)}}
        \end{axis}
    \end{tikzpicture}
    \caption{\textbf{Score domain and query budget.}
    Mean TPR@1\%FPR over 8 DDPM checkpoints at matched query budgets. Both TSC scores use the same temporal stencil.}
    \label{fig:ablation-score-query}
\end{minipage}

\par\medskip
\Needspace{18\baselineskip}
\begingroup
\setlength{\columnsep}{0.02\textwidth}
\setlength{\intextsep}{3pt}
\begin{wraptable}{r}{0.58\textwidth}
    \centering
    \captionsetup{justification=ablationjustified,singlelinecheck=false,skip=5pt}
    \caption{\textbf{Temporal reference filters.}
    Mean AUC and TPR@1\%FPR across four datasets trained over DDPM
    settings. The full Gaussian uses 45 levels and is unavailable at
    $q=15$.}
    \label{tab:filter-selection}
    \small
    \setlength{\tabcolsep}{1.8pt}
    \renewcommand{\arraystretch}{1.12}
    \begin{tabular*}{\linewidth}{@{\extracolsep{\fill}}llrrrr@{}}
        \toprule
        & & \multicolumn{2}{c}{$q=15$} & \multicolumn{2}{c}{$q=45$} \\
        \cmidrule(lr){3-4}\cmidrule(l){5-6}
        Filter & Setting & AUC & TPR & AUC & TPR \\
        \midrule
        \textcolor{tscTeal}{\textbf{TSC (endpoints)}} & $\Delta=2.5,\ k=1$
        & \textbf{87.06} & \textbf{35.30}
        & \textbf{88.11} & \textbf{39.26} \\
        Even & $\Delta=2.5,\ k=2$
        & \underline{83.96} & \underline{26.43} & \underline{86.84} & \underline{34.84} \\
        \addlinespace[1pt]
        Gaussian (full) & $\sigma=3$
        & -- & -- & 81.21 & 20.57 \\
        Gaussian (window) & $k=7,\ \sigma=3$
        & 71.84 & 5.72 & 79.53 & 11.28 \\
        Reverse window & $k=7,\ \sigma=0.5$
        & 80.58 & 16.26 & 85.65 & 27.95 \\
        \bottomrule
    \end{tabular*}
\end{wraptable}
\noindent\phantomsection\label{abl:filter-selection}%
\textbf{Filter selection.}
At $\lambda^\star=0.75$, we compare endpoint references with evenly
spaced, Gaussian, and reverse-window filters, the latter assigning
greater weight to distant levels. Here, $k$ denotes the number of sampled levels on each side of
the center, giving $2k+1$ queried levels in total. TSC uses $k=1$,
with references at $\lambda^\star\pm\Delta$. Definitions and search grids appear in
Appendix~\ref{app:filter-selection}.

Endpoints achieve the highest mean AUC and TPR@1\%FPR at both query
budgets (Table~\ref{tab:filter-selection}). The reverse window improves
over the center-weighted Gaussian window but remains below endpoints.
At $q=45$, repeated noise draws at fixed endpoints outperform the
tested alternatives using additional temporal levels, supporting
concentrating queries on the outer references.
\par\endgroup

\par\medskip
\Needspace{20\baselineskip}
\begingroup
\setlength{\columnsep}{0.02\textwidth}
\setlength{\intextsep}{3pt}
\begin{wraptable}{r}{0.38\textwidth}
    \centering
    \captionsetup{justification=ablationjustified,singlelinecheck=false,skip=5pt}
    \caption{\textbf{Temporal reference symmetry.} At $q=15$, mean AUC and TPR@1\%FPR across four datasets trained over DDPM
    settings. $(a,b)$ denotes nominal log-SNR offsets.}
    \label{tab:symmetry-ablation}
    \small
    \setlength{\tabcolsep}{2pt}
    \renewcommand{\arraystretch}{1.12}
    \begin{tabular*}{\linewidth}{@{\extracolsep{\fill}}llrr@{}}
        \toprule
        Reference & $(a,b)$ & AUC & TPR \\
        \midrule
        \textcolor{tscTeal}{Symmetric} & $(2.5,2.5)$ & \underline{87.06} & \textbf{35.30} \\
        Left only & $(2.5,-)$ & 78.15 & 23.52 \\
        Right only & $(-,2.5)$ & 82.13 & 6.47 \\
        Unequal & $(1.5,3.5)$ & \textbf{88.77} & \underline{30.94} \\
        Unequal & $(3.5,1.5)$ & 80.64 & 26.56 \\
        Weighted & $(1.5,3.5)$ & 86.24 & 29.11 \\
        Weighted & $(3.5,1.5)$ & 83.62 & 26.58 \\
        \bottomrule
    \end{tabular*}
\end{wraptable}

\noindent\phantomsection\label{abl:reference-symmetry}%
\textbf{Reference symmetry.}
At $\lambda^\star=0.75$ and $q=15$, we compare our symmetric stencil against single-sided references and unequal windows spanning $a+b=5$. To maintain a fixed query budget, single-reference scores receive proportionally more draws per evaluated point. As shown in Table~\ref{tab:symmetry-ablation}, symmetry achieves the highest mean TPR@1\%FPR, outperforming all one-sided and asymmetric baselines. While an equal-weight $(1.5,3.5)$ window yields a slightly higher overall AUC, the symmetric contrast strictly dominates in the low-FPR regime - the critical operating requirement for practical auditing. Furthermore, interpolation weighting fails to consistently improve asymmetric windows, empirically validating our symmetric default. Full theoretical diagnostics and per-dataset results are provided in Appendix~\ref{app:symmetry-ablation}.
\par\endgroup

%% file: 06_conclusion.tex
\section{Conclusion}
We introduced difficulty calibration across two time axes to mitigate the image-difficulty confound in diffusion MIAs. \emph{Reference-Model Calibration (RMC)} calibrates across \emph{training time} using a pretrained checkpoint, while \emph{Temporal Self-Calibration (TSC)} calibrates across \emph{diffusion time} using the image's own denoising trajectory. Experiments across diverse datasets, architectures, and training objectives demonstrate substantial gains in strict low-FPR membership detection. RMC enables near-perfect uncaptioned auditing of fine-tuned data, and TSC frequently doubles the TPR@1\%FPR of the strongest trajectory-based baselines in fully unconstrained settings.

\paragraph{Limitations and future work.}
Although our methods use only forward predictions and do not require
gradients, they assume access to denoising outputs for candidate
images at selected noise levels. This access may be unavailable
through APIs that expose only final generated images, limiting
direct application in such settings.

RMC additionally assumes access to the checkpoint from which the
audited model was fine-tuned. Its applicability is therefore
restricted when this checkpoint is unavailable or the model's
training lineage is unknown. Future work could investigate whether
related public checkpoints or ensembles of reference models can
provide effective difficulty calibration in these cases.

RMC measures loss changes between checkpoints and does
not separately identify membership in the reference model's
pretraining corpus. Studying overlap between pretraining and
fine-tuning data, and extending calibration across successive
checkpoints, could clarify how membership evidence evolves
across training stages.

%% file: 06_statements.tex
\subsection*{Ethics statement}
Membership inference has both privacy risks and legitimate auditing uses.
Improved attacks could be misused to infer an individual's participation in
sensitive training data. At the same time, these methods can help individuals
investigate whether their personal data were used to train or fine-tune
diffusion models, and can support audits of candidate training-data sources
for publicly released models, including those distributed through platforms
such as Hugging Face. Such audits can help assess claims about dataset use
and improve transparency around model provenance. Membership scores provide
statistical evidence about candidate examples: a negative result does not
prove that an example was excluded from training, and an audit does not
identify a model's complete training dataset. We advocate using these methods
for authorized privacy auditing and accountability, with conclusions that
account for false positives and false negatives.

%% file: 07_appendix.tex
\appendix

\section{Prediction and Loss Parameterizations}
\label{app:pred-loss-spaces}

As discussed in Section~\ref{sec:prelimineries}, a diffusion or flow-matching model can be parameterized to predict the clean data $x$, the noise $\epsilon$, or the velocity $v$, and its training objective can be evaluated in any of these three spaces. Because these quantities are deterministically linked by the forward process, a prediction in one space can be algebraically converted to any other space before computing the loss. 

Table~\ref{tab:pred-loss-spaces} details the exact conversion equations for all nine combinations under the linear interpolation schedule $z_t = (1-t)\epsilon + tx$, following \citet{li2026back}.

\begin{table}[ht]
\centering
\caption{Prediction and loss parameterizations for the linear flow schedule. Columns specify the direct network prediction space, while rows specify the space in which its squared error is measured.}
\label{tab:pred-loss-spaces}
\vspace{1ex}
\footnotesize
\renewcommand{\arraystretch}{1.6} 
\begin{tabular}{@{} r @{\hspace{0.5em}} l@{}c@{}l c c c @{}} 
\toprule
\multicolumn{4}{@{}l}{} 
& \textbf{$x$-prediction}
& \textbf{$\epsilon$-prediction}
& \textbf{$v$-prediction} \\
\multicolumn{4}{@{}l}{}
& $x_\theta := f_\theta(z_t,t,c)$
& $\epsilon_\theta := f_\theta(z_t,t,c)$
& $v_\theta := f_\theta(z_t,t,c)$ \\
\midrule
\textbf{$x$-loss:} & $\mathbb{E}\|$ & $x_\theta-x$ & $\|_2^2$
& $x_\theta$
& $x_\theta = \frac{z_t-(1-t)\epsilon_\theta}{t}$
& $x_\theta = z_t+(1-t)v_\theta$ \\
\textbf{$\epsilon$-loss:} & $\mathbb{E}\|$ & $\epsilon_\theta-\epsilon$ & $\|_2^2$
& $\epsilon_\theta = \frac{z_t-tx_\theta}{1-t}$
& $\epsilon_\theta$
& $\epsilon_\theta = z_t-tv_\theta$ \\
\textbf{$v$-loss:} & $\mathbb{E}\|$ & $v_\theta-v$ & $\|_2^2$
& $v_\theta = \frac{x_\theta-z_t}{1-t}$
& $v_\theta = \frac{z_t-\epsilon_\theta}{t}$
& $v_\theta$ \\
\bottomrule
\end{tabular}
\end{table}

\section{Experimental Details}
\label{app:experimental-details}

\paragraph{Common protocol.}
For all membership-inference experiments, each target model is trained only on a designated
member split, and attacks are evaluated by distinguishing this member split from a disjoint
hold-out split. Unless stated otherwise, images are normalized to $[-1,1]$, models are
trained with Adam, and we evaluate the final EMA checkpoint. 

\paragraph{DDPM training.}
For the unconditional DDPM experiments, we use a UNet backbone with base width $128$,
channel multipliers $(1,2,2,2)$, two residual blocks per resolution, dropout $0.1$, and
attention at the $16\times16$ resolution. We train $\epsilon$-prediction DDPMs with
$T=1000$ diffusion steps, a linear noise schedule from $\beta_1=10^{-4}$ to
$\beta_T=0.02$, fixed-large variance, learning rate $2\cdot10^{-4}$, gradient clipping
at $1.0$, EMA decay $0.9999$, and batch size $128$. All four datasets below use random
horizontal flips.

\begin{table}[h]
\centering
\caption{DDPM training settings. All runs use $32\times32$ resolution, batch size $128$,
linear noise schedule, $\epsilon$-prediction, and random horizontal flips.}
\label{tab:ddpm-training-settings}
\begin{tabular}{lcccccc}
\toprule
Dataset & Resolution & Member & Hold-out & Aug. & Steps & Epochs \\
\midrule
CIFAR-10   & $32\times32$ & 25k & 25k & HFlip & 800k  & $\approx 4$k \\
CIFAR-100   & $32\times32$ & 25k & 25k & HFlip & 800k  & $\approx 4$k \\
STL-10-U   & $32\times32$ & 50k & 50k & HFlip & 1.6M  & $\approx 4$k \\
Tiny ImageNet & $32\times32$ & 50k & 50k & HFlip & 1.6M & $\approx 4$k \\
\bottomrule
\end{tabular}
\end{table}

\paragraph{Stable Diffusion fine-tuning.}
For text-to-image experiments, we fine-tune Stable Diffusion v1.4 at $512\times512$
resolution. We fine-tune the UNet while keeping the VAE and text encoder fixed, use
mixed precision, gradient checkpointing, EMA, a constant learning-rate schedule, no warmup,
gradient clipping at $1.0$, batch size $1$, gradient accumulation $4$, and learning rate
$10^{-5}$. Unless stated otherwise, all runs use random cropping.

\begin{table}[h]
\centering
\caption{Stable Diffusion v1.4 overfitting-setting fine-tuning details (Table~\ref{tab:sd-finetuned-results}) settings.}
\label{tab:sd-overfit-training-settings}
\begin{tabular}{lcccc}
\toprule
Dataset & Train & Hold-out & Steps & Aug. \\
\midrule
Pok\'emon & 416  & 417  & 7.5k  & Crop \\
MS-COCO   & 2.5k & 2.5k & 150k & Crop \\
Flickr30k & 2.5k & 2.5k & 150k & Crop \\
\bottomrule
\end{tabular}
\end{table}

\begin{table}[h]
\centering
\caption{Stable Diffusion v1.4 real-world-setting fine-tuning details (Table~\ref{tab:reference-calibration-real}) settings.}
\label{tab:sd-real-training-settings}
\begin{tabular}{lcccc}
\toprule
Dataset & Train & Hold-out & Steps & Aug. \\
\midrule
Pok\'emon & 416  & 417  & 7.5k & Crop+HFlip \\
MS-COCO   & 2.5k & 2.5k & 50k  & Crop+HFlip \\
Flickr30k & 10k  & 10k  & 200k & Crop+HFlip \\
\bottomrule
\end{tabular}
\end{table}

\paragraph{Caption data.}
For Pok\'emon, we use image--caption pairs from the Pok\'emon BLIP-caption dataset. For
MS-COCO and Flickr30k, we use the corresponding image-caption annotations from the
constructed train/hold-out splits. Unless stated otherwise, each image is
paired with its staged caption metadata and no caption dropout is applied.

\paragraph{DDPM++}
In addition to the DDPM models, we evaluate a DDPM++ backbone implemented with EDM~\citep{karras2022elucidating} base code ddpmpp over
CIFAR-100 and switching the loss for the 9 different linear flow matching prediction/loss pairs. This model is trained over the same settings as stated in Table~\ref{tab:ddpm-training-settings} except no data augmentation and for $2000$ epochs. We use this setting to isolate the effect of
the native prediction head and the training-loss target.
Specifically, the architecture, data split, optimizer, and training budget are fixed, while
we vary the prediction/loss variants reported in the corresponding tables.

\subsection{Temporal-Filter Selection}
\label{app:filter-selection}

The filter ablation evaluates 45 log-SNR levels from $-5.5$ to $5.5$ at
intervals of $0.25$, with the center fixed at $\lambda^\star=0.75$.
The endpoint stencil is fixed at $\Delta=2.5$, $k=1$, matching the main
DDPM experiments. Even filters search
$\Delta\in\{0.5,1.0,\ldots,4.0\}$ and use $k\in\{2,\ldots,7\}$.
Gaussian bandwidths are selected from
$\sigma\in\{0.5,1.0,1.5,2.0,3.0\}$. Windowed Gaussian filters use
$k\in\{1,\ldots,7\}$ nearest levels per side. Reverse windows use the
same support construction with weights increasing away from the center,
and search $k\in\{2,\ldots,7\}$.

All restricted supports include the central level and therefore contain
$m=2k+1$ levels. Each support level is evaluated using
$\lfloor q/m\rfloor$ independent noise draws. The even and windowed
Gaussian families are selected under a budget of $q=15$. The full
Gaussian necessarily uses all 45 levels and is selected at $q=45$.
For each non-TSC family, one configuration is selected by mean AUC across
the four first-half shadow models. The membership-score direction is fixed
so lower temporal residuals are member-like. Selected supports and weights
are reused on the matching held-out 4K checkpoints at both query budgets.
Results are macro-averaged over CIFAR-10, CIFAR-100,
STL-10-U, and Tiny ImageNet with horizontal flips.

\section{Temporal Reference Symmetry}
\label{app:temporal-symmetry}

We distinguish the removal of a shared loss level from the additional
effect of symmetric reference points. The identities below hold over any
finite window. They require neither a narrow-window approximation nor
the interpretation of the membership feature as pointwise curvature.
We define
\begin{equation}
    \ell_x(u)=\ell_\theta(x,\lambda^\star+u)
\end{equation}
with the stabilized
log-loss defined in Eq.~\eqref{eq:log-loss}.

\paragraph{Shared difficulty and reflection symmetry.}
Under the idealized factorization
$L_\theta(x,\lambda)=H(x)\widetilde L_\theta(x,\lambda)$,
the unstabilized log-loss contains the additive offset $\log H(x)$.
Every contrast whose coefficients sum to zero removes this offset,
including one-sided and asymmetric contrasts. With the numerical
stabilizer $\delta>0$, this correspondence to multiplicative cancellation
is approximate when $\delta$ is negligible relative to all evaluated
losses. Scalar difficulty alone therefore does not single out symmetry.

For any profile, define its even and odd components about the query point:
\begin{equation}
    E_x(u)=\frac{\ell_x(u)+\ell_x(-u)}{2},
    \qquad
    O_x(u)=\frac{\ell_x(u)-\ell_x(-u)}{2}.
    \label{eq:temporal-even-odd}
\end{equation}
The symmetric temporal score is exactly
\begin{equation}
    S_x=\ell_x(0)-E_x(\Delta).
    \label{eq:temporal-even-score}
\end{equation}
Consequently, adding any sample-dependent offset $a_x$ or odd function
$q_x(-u)=-q_x(u)$ to $\ell_x(u)$ leaves $S_x$ unchanged. This removes
antisymmetric profile variation over the entire window, including
nonlinear variation. A deterministic trajectory component shared by all
images contributes only a common score offset and does not affect
rankings. The relevant nuisance is variation between images.

\paragraph{Relation to one-sided references.}
Define $S_{x,+}=\ell_x(0)-\ell_x(\Delta)$ and
$S_{x,-}=\ell_x(0)-\ell_x(-\Delta)$. Then
\begin{equation}
    S_{x,+}=S_x-O_x(\Delta),
    \qquad
    S_{x,-}=S_x+O_x(\Delta).
    \label{eq:temporal-one-sided}
\end{equation}
Thus each one-sided contrast includes the same centered deviation and an
additional odd component with opposite signs. Let $M\in\{0,1\}$ denote
membership, abbreviate $S=S_x$, $S_\pm=S_{x,\pm}$, and
$O=O_x(\Delta)$, and write
$\operatorname{Var}_m$ and $\operatorname{Cov}_m$ for variance and
covariance conditional on $M=m$. Assuming finite second moments,
\begin{equation}
    \operatorname{Var}_m(S_\pm)=\operatorname{Var}_m(S)+\operatorname{Var}_m(O)
      \mp 2\operatorname{Cov}_m(S,O),
      \label{eq:temporal-one-sided-variance}
\end{equation}
\begin{equation}
    \frac{\operatorname{Var}_m(S_+)+\operatorname{Var}_m(S_-)}{2}
    =\operatorname{Var}_m(S)+\operatorname{Var}_m(O).
      \label{eq:temporal-average-variance}
\end{equation}
    
If $O$ has equal conditional means for members and non-members, both
one-sided contrasts have the same mean membership gap as $S$, while
their average conditional variance is larger whenever $O$ varies.
Both individual variances exceed that of $S$ when
$\operatorname{Var}_m(O)>2|\operatorname{Cov}_m(S,O)|$.
These identities explain a possible stability advantage, they do not
by themselves establish an ordering of ROC curves or low-FPR detection.
The identities compare the same per-coordinate loss estimates: since
a one-sided score can use more draws per level under a fixed total query
budget, variance diagnostics require a common number of draws per level,
separately from the comparison at matched total query budgets.

\paragraph{Unequal reference distances.}
Consider references at offsets $-a$ and $b$, with $a,b>0$. The
equal-weight contrast
$U_{x,a,b}=\ell_x(0)-[\ell_x(-a)+\ell_x(b)]/2$
retains $\beta_x(a-b)/2$ from an affine sample-dependent component
$\alpha_x+\beta_xu$. Interpolation at the query point instead gives
\begin{equation}
    T_{x,a,b}=\ell_x(0)
      -\frac{b}{a+b}\ell_x(-a)-\frac{a}{a+b}\ell_x(b),
    \label{eq:temporal-asymmetric-weighted}
\end{equation}
which cancels every such affine component exactly. Applying
Eq.~\eqref{eq:temporal-even-odd} gives the finite-window identity
\begin{align}
    T_{x,a,b}
    &=\ell_x(0)-\frac{bE_x(a)+aE_x(b)}{a+b}
      +R_{x,a,b},\nonumber\\
    R_{x,a,b}
    &=\frac{bO_x(a)-aO_x(b)}{a+b}
      =\frac{ab}{a+b}
       \left[\frac{O_x(a)}{a}-\frac{O_x(b)}{b}\right]
    \label{eq:temporal-asymmetric-odd}
\end{align}
Weighting therefore removes a linear odd component but generally retains
nonlinear odd variation. For example, $O_x(u)=\beta_xu+c_xu^3$
contributes $R_{x,a,b}=c_xab(a-b)$, whereas $a=b$ cancels every odd
component. Unequal distances also sample the even component at different
radii, which can change the retained membership contrast. Conversely,
for an exactly quadratic profile $\ell_x(u)=A_x+B_xu+C_xu^2$,
$T_{x,a,b}=-abC_x$ and $S_x=-\Delta^2C_x$, so these two scores have
identical rankings for fixed positive $a,b,\Delta$.

\paragraph{Scope of the explanation.}
The benefit of symmetry depends on the removed odd variation being
primarily nuisance and on the centered contrast retaining membership
information. An odd membership component would also be removed, and
sample-dependent even variation remains. Symmetry is therefore a useful
invariance, not a universal optimality claim or a guarantee that all
image difficulty is eliminated. In particular, smoothness of difficulty
alone does not ensure its cancellation over a wide window. This account
is compatible with a broad membership feature and does not identify a
unique latent difficulty factor from observed losses.

These statements refer to symmetry in the evaluated log-SNR coordinates.
Mapping requested levels to a discrete diffusion schedule can yield
slightly unequal realized distances. In that case, $a$ and $b$ in
Eq.~\eqref{eq:temporal-asymmetric-weighted} should be computed from the
realized coordinate. Nominally equal spacing alone does not imply exact
odd-component cancellation.

\subsection{Symmetry Ablation Protocol}
\label{app:symmetry-ablation}

We use the same four held-out DDPM checkpoints at 4K epochs as the
filter-selection ablation: CIFAR-10, CIFAR-100,
STL-10-U, and Tiny ImageNet with horizontal flips. 
The query center is fixed at the deployed $\lambda^\star=0.75$.
The symmetric references have nominal distances $(a,b)=(2.5,2.5)$.
We compare each single reference separately and two predeclared unequal
placements, $(a,b)=(1.5,3.5)$ and $(3.5,1.5)$, using either equal
weights or the interpolation weights in
Eq.~\eqref{eq:temporal-asymmetric-weighted}. Thus all two-reference
contrasts have the same nominal span $a+b=5$.
Neither the placement nor the score direction is selected using
held-out membership labels, smaller residuals are always member-like.

At each noise level, we compute
\begin{equation}
    \widehat\ell_{x,K}(\lambda)
    =\log\!\left(\frac{1}{K}\sum_{j=1}^{K}
             L_\theta(x,\lambda,\epsilon_j)+10^{-12}\right).
    \label{eq:symmetry-empirical-log-loss}
\end{equation}
At budgets
$q\in\{15,45\}$, three-point contrasts use $K=q/3$ draws at each level.
One-reference contrasts use $\lceil q/2\rceil$ draws at the center and
$\lfloor q/2\rfloor$ at the reference, assigning the remaining query
to the center. Every configuration therefore uses exactly $q$ model
queries. The odd-component diagnostics instead use $q/3$ draws at all
three levels, so the identities in Eq.~\eqref{eq:temporal-one-sided}
compare the same loss estimates.

Requested log-SNR levels are mapped to the nearest available DDPM
timesteps. The nominal symmetric triple $(-1.75,0.75,3.25)$ is realized
as approximately $(-1.752006,0.755986,3.264399)$. Its midpoint differs
from the query level by approximately $2.11\times10^{-4}$.
We retain equal weights for this deployed TSC baseline and use the
realized distances for the unequal interpolation weights.
We compute AUC and TPR@1\%FPR separately for each checkpoint and then
take their unweighted mean, rather than pooling scores across models.

\paragraph{Results by checkpoint.}
Table~\ref{tab:symmetry-per-model} reports the individual results underlying
the main-text averages. Symmetry has the highest mean TPR@1\%FPR across
these four checkpoints at both budgets, but does not dominate every
checkpoint or metric. The equal-weight $(1.5,3.5)$ layout has higher AUC
on all four checkpoints. The effect of interpolation weighting depends
on the orientation and metric, consistent with its changing both nuisance
variation and the membership contrast retained by the references.

\begin{table}[ht]
    \centering
    \caption{\textbf{Reference symmetry by checkpoint.} All checkpoints are held-out DDPMs at 4K epochs. AUC and TPR@1\%FPR are percentages. C100 denotes CIFAR-100 with horizontal flips. Bold and underlined values denote the best and second-best results within each query budget.}
    \label{tab:symmetry-per-model}
    \small
    \setlength{\tabcolsep}{3.2pt}
    \renewcommand{\arraystretch}{1.12}
    \begin{tabular*}{\linewidth}{@{\extracolsep{\fill}}rlrrrrrrrr@{}}
        \toprule
        & & \multicolumn{2}{c}{CIFAR-10} & \multicolumn{2}{c}{C100 hflip} & \multicolumn{2}{c}{STL-10-U} & \multicolumn{2}{c}{Tiny ImageNet} \\
        \cmidrule(lr){3-4}\cmidrule(lr){5-6}\cmidrule(lr){7-8}\cmidrule(l){9-10}
        $q$ & Reference $(a,b)$ & AUC & TPR & AUC & TPR & AUC & TPR & AUC & TPR \\
        \midrule
        15 & \textcolor{tscTeal}{Symmetric} $(2.5,2.5)$ & \underline{95.91} & \textbf{62.31} & 95.85 & \textbf{57.06} & \underline{77.84} & \textbf{9.57} & \underline{78.63} & \underline{12.27} \\
        15 & Left only $(2.5,-)$ & 88.78 & 42.48 & 88.95 & 38.22 & 67.39 & 6.42 & 67.48 & 6.97 \\
        15 & Right only $(-,2.5)$ & 91.70 & 9.76 & 92.11 & 12.09 & 71.34 & 1.62 & 73.37 & 2.41 \\
        15 & Unequal $(1.5,3.5)$ & \textbf{97.27} & \underline{55.60} & \textbf{97.32} & \underline{54.35} & \textbf{80.12} & 6.79 & \textbf{80.38} & 7.02 \\
        15 & Unequal $(3.5,1.5)$ & 90.18 & 44.90 & 90.24 & 40.80 & 70.74 & \underline{8.63} & 71.42 & 11.91 \\
        15 & Weighted $(1.5,3.5)$ & 95.71 & 52.60 & \underline{95.95} & 48.96 & 77.26 & 7.67 & 76.04 & 7.21 \\
        15 & Weighted $(3.5,1.5)$ & 92.13 & 45.10 & 92.00 & 40.31 & 74.32 & 8.62 & 76.05 & \textbf{12.31} \\
        \midrule
        45 & \textcolor{tscTeal}{Symmetric} $(2.5,2.5)$ & 96.45 & \textbf{67.80} & 96.43 & \underline{62.60} & \underline{79.60} & \textbf{12.40} & \underline{79.96} & \textbf{14.26} \\
        45 & Left only $(2.5,-)$ & 89.35 & 44.31 & 89.47 & 40.82 & 68.05 & 6.61 & 68.04 & 7.53 \\
        45 & Right only $(-,2.5)$ & 91.99 & 10.29 & 92.51 & 11.96 & 71.76 & 1.64 & 73.78 & 2.32 \\
        45 & Unequal $(1.5,3.5)$ & \textbf{97.96} & \underline{65.65} & \textbf{97.93} & \textbf{62.88} & \textbf{81.16} & 7.01 & \textbf{82.04} & 7.72 \\
        45 & Unequal $(3.5,1.5)$ & 91.03 & 48.08 & 91.03 & 45.22 & 71.72 & 10.05 & 71.90 & 13.02 \\
        45 & Weighted $(1.5,3.5)$ & \underline{96.89} & 64.97 & \underline{97.06} & 61.85 & 78.59 & 8.79 & 78.08 & 8.68 \\
        45 & Weighted $(3.5,1.5)$ & 93.39 & 51.12 & 93.06 & 46.58 & 76.03 & \underline{10.18} & 76.73 & \underline{13.90} \\
        \bottomrule
    \end{tabular*}
\end{table}

\paragraph{Variation removed by the symmetric contrast.}
Table~\ref{tab:symmetry-components} compares the center log-loss, the
symmetric contrast $S$, the endpoint difference $O$, and the two
one-sided contrasts, using the same $K=5$ draws at every point.
For each checkpoint we compute the non-member minus member mean gap
and the pooled within-class standard deviation
$\sqrt{[\operatorname{Var}(S\mid M=0)+\operatorname{Var}(S\mid M=1)]/2}$,
with analogous definitions for the other components, and then average
over checkpoints. The left-only contrast retains a larger mean gap
than $S$ but has substantially greater spread. The smaller separation
of $O$ relative to its spread is consistent with suppression of
difficulty-related variation, without establishing that $O$ is pure
nuisance. Its mean AUC is above chance, and covariance with $S$ need
not vanish. These are descriptive diagnostics at common per-point
precision. The one-sided rows here use ten queries, whereas the
symmetric contrast uses fifteen. They are distinct from the matched
budget comparisons in Table~\ref{tab:symmetry-ablation}.

\begin{table}[ht]
    \centering
    \caption{\textbf{Components of the temporal contrast.} Four-checkpoint means at $K=5$ shared draws per point. Gaps and standard deviations are in log-loss units.}
    \label{tab:symmetry-components}
    \small
    \begin{tabular}{@{}lrrr@{}}
        \toprule
        Component & Mean gap & Pooled SD & AUC \\
        \midrule
        Center $\ell_x(0)$ & 0.3885 & 0.4119 & 74.26 \\
        Symmetric $S$ & 0.2566 & 0.1366 & 87.06 \\
        Endpoint difference $O$ & 0.0371 & 0.1552 & 57.14 \\
        Left reference $S+O$ & 0.2937 & 0.2464 & 77.80 \\
        Right reference $S-O$ & 0.2195 & 0.1580 & 81.84 \\
        \bottomrule
    \end{tabular}
\end{table}

\section{Prediction-Loss Analysis}
\label{app:pred-loss-analysis}

This section expands the analysis in Section~\ref{sec:loss_temporal_localization}.
We use $p\in\{x,v,\epsilon\}$ for the native prediction head and
$s\in\{x,v,\epsilon\}$ for the training-loss target. The symbol
$\mathcal{L}$ denotes squared error, while the calibration method uses $\ell$ for
log-loss. $q$ remains the number of model queries per candidate.

\subsection{Conversions and Objective Weighting}
\label{app:fm_parameterization_conversion}

For the linear flow $z_t=tx+(1-t)\epsilon$, with $v=x-\epsilon$ and
$t\in(0,1)$, the identities
\begin{equation}
    x=z_t+(1-t)v, \qquad \epsilon=z_t-tv
    \label{eq:flow_identities}
\end{equation}
give the conversions in Table~\ref{tab:pred-loss-spaces}. In particular,
each native prediction induces a velocity prediction:
\begin{equation}
    \widehat v_\theta^{(x)}=\frac{\widehat x_\theta-z_t}{1-t},
    \qquad \widehat v_\theta^{(v)}=\widehat v_\theta,
    \qquad \widehat v_\theta^{(\epsilon)}
        =\frac{z_t-\widehat\epsilon_\theta}{t}.
    \label{eq:implied_velocity_triplet}
\end{equation}
Fix one model $\theta$ with native head $p$, and convert its output to each
loss space. The resulting residuals satisfy
\begin{equation}
    \delta_v^{(p)}=\widehat v_\theta^{(p)}-v, \qquad
    \delta_x^{(p)}=(1-t)\delta_v^{(p)}, \qquad
    \delta_\epsilon^{(p)}=-t\delta_v^{(p)}.
    \label{eq:residual_conversion}
\end{equation}
All three residuals here come from the same model output. This identity
makes no assertion that separately trained prediction heads produce the
same errors. For example, the expanded $x$-loss is
\begin{equation}
    \begin{aligned}
        \mathcal{L}_{p,x}(x,\epsilon,t,\theta)
        &=\left\|\bigl(z_t+(1-t)\widehat v_\theta^{(p)}\bigr)
                  -\bigl(z_t+(1-t)v\bigr)\right\|_2^2 \\
        &=(1-t)^2\left\|\delta_v^{(p)}\right\|_2^2.
    \end{aligned}
    \label{eq:app-x-loss-expansion}
\end{equation}
Squaring the other residuals gives the common-coordinate factorization
used in Section~\ref{sec:loss_temporal_localization}:
\begin{equation}
    \mathcal{L}_{p,s}(x,\epsilon,t,\theta)
        =w_s(t)\left\|\delta_v^{(p)}\right\|_2^2,
    \qquad (w_x(t),w_v(t),w_\epsilon(t))=((1-t)^2,1,t^2).
    \label{eq:loss_weight_triplet}
\end{equation}
For timestep sampling density $\rho(t)$, the population objective is
therefore
\begin{equation}
    \mathcal{R}_{p,s}(\theta)
    =\int_0^1 \rho(t)w_s(t)
        \mathbb{E}_{x,\epsilon}
        \left[\left\|\delta_v^{(p)}\right\|_2^2\right]dt.
    \label{eq:app-population-risk}
\end{equation}
The weight $w_s$ is determined by the loss target, but the residual depends
on the fitted model. Native prediction heads also have different output
coordinates and parameter Jacobians. Thus, the common-coordinate
factorization does not imply identical optimization or learned models.
The analysis concerns interior timesteps, where the conversions are
well-defined, before any numerical clamping at the endpoints.

For a fixed model and fixed interior timestep, evaluating its output in
different loss spaces multiplies the squared error by a positive factor
shared by all images. Averaging over the same noise draws preserves this
relation, so the single-timestep rankings, ROC curves, and AUC are
unchanged. This invariance concerns changing the evaluation space of one
model. The nine configurations in Table~\ref{tab:pred-loss-results}
are separately trained models with different prediction/loss choices.

\subsection{Temporal-Weight Barycenters and Their Ordering}
\label{app:barycenter_proof}

For a common timestep sampling distribution $\rho$, define the barycenter
of the explicit temporal weights as
\begin{equation}
    c_s=\frac{\mathbb{E}_{T\sim\rho}[T w_s(T)]}
                 {\mathbb{E}_{T\sim\rho}[w_s(T)]}.
    \label{eq:weight_mean}
\end{equation}
This quantity summarizes the explicit weights $\rho(t)w_s(t)$,
independently of the model-dependent residual in
Eq.~(\ref{eq:app-population-risk}). For a sampling density $\rho(t)$,
the corresponding normalized density is
\begin{equation}
    \pi_s(t)=\frac{\rho(t)w_s(t)}
                   {\int_0^1\rho(u)w_s(u)\,du},
    \qquad c_s=\int_0^1 t\pi_s(t)\,dt.
    \label{eq:normalized_target_weight}
\end{equation}
\paragraph{Uniform timestep sampling.}
For $\rho(t)=1$, the normalization constants are
$\int_0^1(1-t)^2dt=\int_0^1t^2dt=1/3$, giving
$\pi_x(t)=3(1-t)^2$, $\pi_v(t)=1$, and $\pi_\epsilon(t)=3t^2$.
Their barycenters are
\begin{equation}
    \begin{aligned}
        c_x&=\frac{\int_0^1 t(1-t)^2\,dt}
                   {\int_0^1(1-t)^2\,dt}
             =\frac{1/12}{1/3}=\frac14, \\
        c_v&=\int_0^1 t\,dt=\frac12, \\
        c_\epsilon&=\frac{\int_0^1t^3\,dt}
                          {\int_0^1t^2\,dt}
             =\frac{1/4}{1/3}=\frac34.
    \end{aligned}
    \label{eq:uniform_center_order}
\end{equation}
Thus the centers of the explicit weights move from the noisy toward the
clean end as the training loss changes from $x$ to $v$ to $\epsilon$.
The factor $t$ in these integrals measures the location of each weight,
it is not an additional factor in the training objective.

\paragraph{General timestep sampling.}
Let $T\sim\rho$ have nonzero variance and take values in $(0,1)$.
Since $w_v=1$, $c_v=\mathbb{E}[T]$, and
\begin{equation}
    c_s-c_v
    =\frac{\operatorname{Cov}(T,w_s(T))}
           {\mathbb{E}[w_s(T)]}.
    \label{eq:app-barycenter-covariance}
\end{equation}
For an independent copy $T'$ of $T$ and any function $g$ used below,
\begin{equation}
    2\operatorname{Cov}(T,g(T))
    =\mathbb{E}\!\left[(T-T')\bigl(g(T)-g(T')\bigr)\right].
    \label{eq:app-monotone-covariance}
\end{equation}
For $g(t)=(1-t)^2$, which is strictly decreasing on $(0,1)$, the integrand is negative whenever $T\ne T'$. Conversely, for $g(t)=t^2$, which is strictly increasing, the integrand is positive. Hence
\begin{equation}
    c_x<c_v<c_\epsilon.
    \label{eq:general_center_order}
\end{equation}
For discrete sampling, replacing the integrals by sums gives the same
proof. A distribution concentrated at one interior timestep instead gives
equality throughout.

These centers describe only the explicit objective weighting. They are
neither the centers of the actual training-error contribution nor the
maxima of a membership-inference curve. The latter also depend on the
learned errors, optimization, and sample information at each noise level.
The ordering above is therefore a statement about explicit objective weights,
not a theorem about membership peaks. It provides an interpretation of the
empirical shifts examined below.

\subsection{Empirical Localization of Membership Leakage}
\label{app:loss_auc_curves}

\begin{figure}[t]
    \centering
    \includegraphics[width=\linewidth]{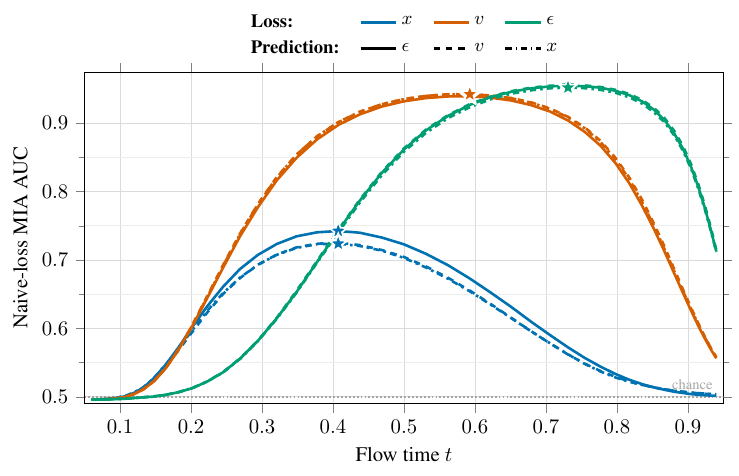}
    \caption{\textbf{Membership localization across training losses.}
    Single-timestep absolute-loss AUC on the nine first-half CIFAR-100
    DDPM++ models without augmentation at epoch $2000$.
    Color denotes training-loss target $s$ and line style denotes native
    prediction $p$. Losses are averaged over $50$ noise draws per image,
    shared across noise levels. Lower loss indicates membership at every
    level. Stars mark single-timestep AUC maxima. The peaks occur at
    $\lambda=-0.75$, $0.75$, and $2.00$ for $x$-, $v$-, and
    $\epsilon$-loss, respectively.}
    \label{fig:edm_loss_localization}
\end{figure}

Figure~\ref{fig:edm_loss_localization} uses the nine CIFAR-100
DDPM++ models trained without augmentation for $2000$ epochs.
For each model, we evaluate the single-timestep absolute reconstruction loss in its training
loss space on the grid
$\mathcal{G}=\{-5.5,-5.25,\ldots,5.5\}$ of $45$ log-SNR values.
Each curve uses $25{,}000$ members and $25{,}000$ non-members, with
$K=50$ deterministic Gaussian noise draws per image reused across noise
levels. Losses are averaged over draws before taking the logarithm,
this monotone transformation leaves single-timestep rankings unchanged.

Writing $\theta_{p,s}$ for the model trained with head $p$ and loss target
$s$, the underlying lower-loss membership score is
\begin{equation}
    S_{p,s}(x,\lambda)
    =-\frac{1}{K}\sum_{k=1}^{K}
       \mathcal{L}_{p,s}\bigl(x,\epsilon_k,t(\lambda),\theta_{p,s}\bigr),
    \qquad t(\lambda)=\frac{1}{1+e^{-\lambda/2}}.
    \label{eq:single_t_mia_score}
\end{equation}
Let $A_{p,s}(\lambda)$ denote the ROC AUC of $S_{p,s}$, with members as
the positive class. This fixes lower reconstruction loss as member-like at
every noise level. The empirical leakage peak is
\begin{equation}
    \lambda^{\mathrm{peak}}_{p,s}
    \in\operatorname*{arg\,max}_{\lambda\in\mathcal{G}}
        A_{p,s}(\lambda).
    \label{eq:optimal_mia_t}
\end{equation}
All three native heads attain the same discrete peak within each training
loss: $\lambda^{\mathrm{peak}}=-0.75$, $0.75$, and $2.00$ for
$x$-, $v$-, and $\epsilon$-loss, respectively. These correspond to flow
times $0.407$, $0.593$, and $0.731$. In this experiment, the membership
signal therefore shifts toward higher log-SNR as the training loss changes
from $x$ to $v$ to $\epsilon$, consistent with the explicit-weight ordering.
The observed peaks remain interior and differ from the barycenters. Their
locations are empirical properties of the trained models and evaluation.

The empirical peak $\lambda^{\mathrm{peak}}_{p,s}$ is measured separately
for each model using the single-timestep score, it is distinct from the
shadow-selected TSC center, which optimizes a three-point contrast.
Appendix~\ref{app:full-pred-loss-results} reports the full attack results
across all nine prediction/loss configurations, with and without augmentation.

\section{Additional Experiments}
\label{app:additional-experiments}

\raggedbottom

Unless stated otherwise, all results in this section use the held-out second
evaluation split, with operating points selected on the corresponding first-half
shadow checkpoints. TSC uses the endpoint stencil with $\Delta=2.5$ and five
noise draws per trajectory point ($q=15$). We do not apply frequency filtering
to any attack. AUC, ASR, and TPR@1\%FPR are reported as percentages. ASR uses
the oracle threshold on the target evaluation split.

\subsection{Earlier DDPM Checkpoints}
\label{app:ddpm-2k}

Tables~\ref{tab:app-ddpm-2k} and~\ref{tab:app-ddpm-2k-stl-tinyin} complement the 4K-checkpoint results in the main
paper with the corresponding 2K checkpoints. The shorter training horizon
weakens membership leakage for every attack, particularly on STL-10-U and
Tiny ImageNet, yet TSC remains the strongest method under every metric and
dataset.

\par\medskip
\noindent\begin{minipage}{\linewidth}
    \centering
    \captionsetup{skip=5pt}
    \captionof{table}{\textbf{Membership inference at 2K DDPM checkpoints: CIFAR datasets.} Standard training setting, parentheses give model queries per candidate.}
    \label{tab:app-ddpm-2k}
    \begingroup
    \small
    \setlength{\tabcolsep}{3.2pt}
    \renewcommand{\arraystretch}{1.12}
    \begin{tabular*}{\linewidth}{@{\extracolsep{\fill}}llrrrrrr@{}}
        \toprule
        & & \multicolumn{3}{c}{CIFAR-10} & \multicolumn{3}{c}{CIFAR-100} \\
        \cmidrule(lr){3-5}\cmidrule(lr){6-8}
        \multicolumn{2}{l}{Method} & AUC & ASR & TPR & AUC & ASR & TPR \\
        \midrule
        Naive & ($q=20$) & 70.70 & 65.44 & 2.75 & 69.31 & 64.18 & 4.19 \\
        SecMI & ($q=12$) & 74.15 & 68.14 & 3.61 & 72.74 & 66.89 & 4.00 \\
        PIA & ($q=2$) & \underline{77.09} & \underline{70.75} & 6.07 & \underline{75.82} & \underline{69.48} & 6.55 \\
        PIAN & ($q=2$) & 74.24 & 69.18 & \underline{8.08} & 72.75 & 68.07 & \underline{7.74} \\
        \midrule
        TSC & ($q=15$) & \textbf{85.34} & \textbf{78.12} & \textbf{20.61} & \textbf{84.97} & \textbf{77.72} & \textbf{18.24} \\
        \bottomrule
    \end{tabular*}
    \endgroup
    \medskip
    \captionof{table}{\textbf{Membership inference at 2K DDPM checkpoints: STL-10-U and Tiny ImageNet.} Same protocol as Table~\ref{tab:app-ddpm-2k}.}
    \label{tab:app-ddpm-2k-stl-tinyin}
    \begingroup
    \small
    \setlength{\tabcolsep}{3.2pt}
    \renewcommand{\arraystretch}{1.12}
    \begin{tabular*}{\linewidth}{@{\extracolsep{\fill}}llrrrrrr@{}}
        \toprule
        & & \multicolumn{3}{c}{STL-10-U} & \multicolumn{3}{c}{Tiny ImageNet} \\
        \cmidrule(lr){3-5}\cmidrule(lr){6-8}
        \multicolumn{2}{l}{Method} & AUC & ASR & TPR & AUC & ASR & TPR \\
        \midrule
        Naive & ($q=20$) & 56.52 & 54.82 & 1.50 & 55.85 & 54.30 & 1.57 \\
        SecMI & ($q=12$) & \underline{59.28} & \underline{56.83} & 1.63 & 56.85 & 54.95 & 1.52 \\
        PIA & ($q=2$) & 59.15 & 56.68 & 1.84 & \underline{57.51} & \underline{55.60} & 1.84 \\
        PIAN & ($q=2$) & 57.30 & 55.81 & \underline{1.95} & 56.54 & 55.15 & \underline{2.05} \\
        \midrule
        TSC & ($q=15$) & \textbf{63.70} & \textbf{60.11} & \textbf{3.30} & \textbf{62.07} & \textbf{59.03} & \textbf{2.84} \\
        \bottomrule
    \end{tabular*}
    \endgroup
\end{minipage}
\par\medskip

\subsection{DDPMs Without Horizontal Flips}
\label{app:ddpm-no-hflip}

We additionally evaluate STL-10-U and Tiny ImageNet without horizontal-flip
augmentation at both 2K and 4K epochs. The 4K checkpoints reproduce the
no-flip training regime commonly used in prior diffusion-MIA evaluations.
Tables~\ref{tab:app-ddpm-no-hflip} and~\ref{tab:app-ddpm-no-hflip-4k} show that the advantage of TSC persists at
both training horizons and becomes especially pronounced at low FPR after
longer training.

\par\medskip
\noindent\begin{minipage}{\linewidth}
    \centering
    \captionsetup{skip=5pt}
    \captionof{table}{\textbf{Membership inference without horizontal flips at 2K epochs.} Parentheses give model queries per candidate.}
    \label{tab:app-ddpm-no-hflip}
    \begingroup
    \small
    \setlength{\tabcolsep}{3.2pt}
    \renewcommand{\arraystretch}{1.12}
    \begin{tabular*}{\linewidth}{@{\extracolsep{\fill}}llrrrrrr@{}}
        \toprule
        & & \multicolumn{3}{c}{STL-10-U} & \multicolumn{3}{c}{Tiny ImageNet} \\
        \cmidrule(lr){3-5}\cmidrule(lr){6-8}
        \multicolumn{2}{l}{Method} & AUC & ASR & TPR & AUC & ASR & TPR \\
        \midrule
        Naive & ($q=20$) & 78.72 & 72.20 & 4.88 & 77.89 & 71.17 & 5.69 \\
        SecMI & ($q=12$) & 82.71 & 75.66 & 6.85 & 80.91 & 73.90 & 6.46 \\
        PIA & ($q=2$) & \underline{86.68} & \underline{79.39} & 12.59 & \underline{86.11} & \underline{78.93} & 13.52 \\
        PIAN & ($q=2$) & 83.72 & 77.85 & \underline{15.38} & 83.29 & 77.53 & \underline{20.01} \\
        \midrule
        TSC & ($q=15$) & \textbf{93.39} & \textbf{86.88} & \textbf{38.23} & \textbf{93.28} & \textbf{86.55} & \textbf{45.92} \\
        \bottomrule
    \end{tabular*}
    \endgroup
    \medskip
    \captionof{table}{\textbf{Membership inference without horizontal flips at 4K epochs.} Same protocol as Table~\ref{tab:app-ddpm-no-hflip}.}
    \label{tab:app-ddpm-no-hflip-4k}
    \begingroup
    \small
    \setlength{\tabcolsep}{3.2pt}
    \renewcommand{\arraystretch}{1.12}
    \begin{tabular*}{\linewidth}{@{\extracolsep{\fill}}llrrrrrr@{}}
        \toprule
        & & \multicolumn{3}{c}{STL-10-U} & \multicolumn{3}{c}{Tiny ImageNet} \\
        \cmidrule(lr){3-5}\cmidrule(lr){6-8}
        \multicolumn{2}{l}{Method} & AUC & ASR & TPR & AUC & ASR & TPR \\
        \midrule
        Naive & ($q=20$) & 87.63 & 80.38 & 10.04 & 88.02 & 80.47 & 13.71 \\
        SecMI & ($q=12$) & 90.86 & 83.94 & 13.22 & 90.60 & 83.38 & 14.56 \\
        PIA & ($q=2$) & \underline{93.85} & \underline{87.27} & 29.10 & \underline{94.96} & \underline{88.63} & 40.48 \\
        PIAN & ($q=2$) & 91.64 & 85.81 & \underline{33.99} & 91.98 & 87.00 & \underline{51.89} \\
        \midrule
        TSC & ($q=15$) & \textbf{96.98} & \textbf{91.97} & \textbf{56.91} & \textbf{97.59} & \textbf{92.63} & \textbf{77.07} \\
        \bottomrule
    \end{tabular*}
    \endgroup
\end{minipage}
\par\medskip

\subsection{Conditioning and Diffusion Schedule}
\label{app:conditioning-schedule}

\paragraph{Class conditioning.}
We first score the same class-conditional $\epsilon$-prediction,
$\epsilon$-loss SongUNet checkpoint either with the unconditional condition
or with the ground-truth class label. Each evaluation contains 25K members
and 25K non-members. Access to the class condition strengthens every attack,
while TSC remains best in both settings.

\par\medskip
\noindent\begin{minipage}{\linewidth}
    \centering
    \captionsetup{skip=5pt}
    \captionof{table}{\textbf{Effect of class conditioning on a conditional DDPM++.} The same checkpoint is evaluated with the condition dropped or with the ground-truth class. TPR denotes TPR@1\%FPR.}
    \label{tab:app-songunet-conditioning}
    \begingroup
    \small
    \setlength{\tabcolsep}{3.2pt}
    \renewcommand{\arraystretch}{1.12}
    \begin{tabular*}{\linewidth}{@{\extracolsep{\fill}}llrrrrrr@{}}
        \toprule
        & & \multicolumn{3}{c}{Condition dropped} & \multicolumn{3}{c}{Ground-truth class} \\
        \cmidrule(lr){3-5}\cmidrule(lr){6-8}
        \multicolumn{2}{l}{Method} & AUC & ASR & TPR & AUC & ASR & TPR \\
        \midrule
        Naive & ($q=20$) & 67.93 & 63.17 & 4.74 & 77.83 & 71.05 & 8.16 \\
        SecMI & ($q=12$) & 62.19 & 58.61 & 3.25 & 68.25 & 63.02 & 5.06 \\
        PIA & ($q=2$) & \underline{74.86} & \underline{68.89} & 8.53 & \underline{85.29} & \underline{77.52} & 15.20 \\
        PIAN & ($q=2$) & 70.46 & 66.97 & \underline{8.90} & 82.13 & 76.76 & \underline{18.90} \\
        \midrule
        TSC & ($q=15$) & \textbf{85.06} & \textbf{78.27} & \textbf{17.72} & \textbf{92.60} & \textbf{85.86} & \textbf{24.92} \\
        \bottomrule
    \end{tabular*}
    \endgroup
\end{minipage}
\par\medskip

\paragraph{Noise schedule.}
Table~\ref{tab:app-ddpm-schedules} compares class-conditional CIFAR-100 DDPMs
trained for 2K epochs with cosine and linear noise schedules. Although the
absolute leakage differs substantially between schedules, TSC provides the
strongest separation in both cases.

\par\medskip
\noindent\begin{minipage}{\linewidth}
    \centering
    \captionsetup{skip=5pt}
    \captionof{table}{\textbf{Effect of the DDPM noise schedule.} Class-conditional CIFAR-100 checkpoints are evaluated at 2K epochs. TPR denotes TPR@1\%FPR.}
    \label{tab:app-ddpm-schedules}
    \begingroup
    \small
    \setlength{\tabcolsep}{3.2pt}
    \renewcommand{\arraystretch}{1.12}
    \begin{tabular*}{\linewidth}{@{\extracolsep{\fill}}llrrrrrr@{}}
        \toprule
        & & \multicolumn{3}{c}{Cosine schedule} & \multicolumn{3}{c}{Linear schedule} \\
        \cmidrule(lr){3-5}\cmidrule(lr){6-8}
        \multicolumn{2}{l}{Method} & AUC & ASR & TPR & AUC & ASR & TPR \\
        \midrule
        Naive & ($q=20$) & 89.52 & 82.45 & 14.60 & 81.02 & 73.90 & 7.80 \\
        SecMI & ($q=12$) & 88.76 & 81.11 & 17.08 & 82.74 & 75.49 & 8.69 \\
        PIA & ($q=2$) & \underline{95.16} & \underline{88.99} & 30.86 & \underline{86.40} & \underline{78.88} & 12.80 \\
        PIAN & ($q=2$) & 93.09 & 87.61 & \underline{33.03} & 84.41 & 78.43 & \underline{15.53} \\
        \midrule
        TSC & ($q=15$) & \textbf{98.73} & \textbf{94.91} & \textbf{75.32} & \textbf{95.29} & \textbf{89.08} & \textbf{37.20} \\
        \bottomrule
    \end{tabular*}
    \endgroup
\end{minipage}
\par\medskip

\subsection{Full Prediction--Loss Results and Operating-Point Selection}
\label{app:full-pred-loss-results}

The following tables report AUC, ASR, and TPR@1\%FPR for each training-loss
and native-prediction pair, with and without augmentation. The non-augmented models are trained for $2000$ epochs, while the augmented counterparts are trained for $4000$ epochs. 

Naive uses $20$ noise draws at one level ($q=20$). TSC uses five draws
at each of its three levels ($q=15$), with $\Delta=2.5$ and no frequency
filtering. PIA and PIAN use their two-query scores. SecMI follows its original fixed protocol ($q=12$), so no center is selected for it.
Table~\ref{tab:pred-loss-results} summarizes the non-augmented TPR@1\%FPR results.
The following tables below retain all metrics and the augmented configurations. Each loss uses one consistent point $\lambda^*$ shared across prediction parameterizations and augmentation variants. All metrics are in percentages and TPR denotes TPR@1\%FPR. Augmentations are the default EDM~\citep{karras2022elucidating} augmentations.

\par\medskip
\noindent\begin{minipage}{\linewidth}
    \centering
    \captionsetup{skip=5pt}
    \captionof{table}{\textbf{DDPM++ results under $x$-loss without augmentation.}}
    \label{tab:app-songunet-noaug-xloss}
    \begingroup
    \small
    \setlength{\tabcolsep}{3.2pt}
    \renewcommand{\arraystretch}{1.12}
    \begin{tabular*}{\linewidth}{@{\extracolsep{\fill}}llcrrrrrrrrr@{}}
        \toprule
        & & \multirow{2}{*}{$\lambda^\star$} & \multicolumn{3}{c}{$x$-pred} & \multicolumn{3}{c}{$v$-pred} & \multicolumn{3}{c}{$\epsilon$-pred} \\
        \cmidrule(lr){4-6}\cmidrule(lr){7-9}\cmidrule(lr){10-12}
        \multicolumn{2}{l}{Method} & & AUC & ASR & TPR & AUC & ASR & TPR & AUC & ASR & TPR \\
        \midrule
        Naive & ($q=20$) & $-0.75$ & \underline{72.20} & \underline{66.16} & \underline{6.48} & 71.77 & 65.80 & \underline{6.05} & 73.06 & 66.75 & \underline{6.74} \\
        SecMI & ($q=12$) & fixed & 50.57 & 50.67 & 0.94 & 51.78 & 51.38 & 1.10 & 51.03 & 50.98 & 1.19 \\
        PIA & ($q=2$) & $-0.75$ & 58.52 & 57.03 & 2.27 & \underline{72.84} & \underline{67.11} & 4.07 & \underline{74.00} & \underline{67.87} & 6.22 \\
        PIAN & ($q=2$) & $-0.5$ & 55.88 & 55.17 & 2.41 & 68.38 & 64.15 & 6.00 & 67.49 & 63.63 & 5.69 \\
        \midrule
        TSC & ($q=15$) & $-0.5$ & \textbf{81.96} & \textbf{73.94} & \textbf{30.28} & \textbf{81.34} & \textbf{73.50} & \textbf{28.74} & \textbf{82.60} & \textbf{74.52} & \textbf{31.26} \\
        \bottomrule
    \end{tabular*}
    \endgroup
    \medskip
    \captionof{table}{\textbf{DDPM++ results under $x$-loss with augmentation.}}
    \label{tab:app-songunet-aug-xloss}
    \begingroup
    \small
    \setlength{\tabcolsep}{3.2pt}
    \renewcommand{\arraystretch}{1.12}
    \begin{tabular*}{\linewidth}{@{\extracolsep{\fill}}llcrrrrrrrrr@{}}
        \toprule
        & & \multirow{2}{*}{$\lambda^\star$} & \multicolumn{3}{c}{$x$-pred} & \multicolumn{3}{c}{$v$-pred} & \multicolumn{3}{c}{$\epsilon$-pred} \\
        \cmidrule(lr){4-6}\cmidrule(lr){7-9}\cmidrule(lr){10-12}
        \multicolumn{2}{l}{Method} & & AUC & ASR & TPR & AUC & ASR & TPR & AUC & ASR & TPR \\
        \midrule
        Naive & ($q=20$) & $-0.75$ & \underline{56.70} & \underline{54.85} & \underline{1.96} & 56.85 & 54.95 & \underline{1.85} & 57.47 & 55.48 & \underline{1.98} \\
        SecMI & ($q=12$) & fixed & 50.37 & 50.40 & 1.01 & 50.58 & 50.63 & 1.12 & 50.58 & 50.64 & 1.02 \\
        PIA & ($q=2$) & $-0.75$ & 52.25 & 51.77 & 1.14 & \underline{57.57} & \underline{55.64} & 1.75 & \underline{58.06} & \underline{56.01} & 1.94 \\
        PIAN & ($q=2$) & $-0.5$ & 50.92 & 50.82 & 1.07 & 54.91 & 53.80 & 1.54 & 56.39 & 54.76 & 1.64 \\
        \midrule
        TSC & ($q=15$) & $-0.5$ & \textbf{61.47} & \textbf{58.32} & \textbf{5.71} & \textbf{61.58} & \textbf{58.30} & \textbf{5.82} & \textbf{62.43} & \textbf{59.11} & \textbf{6.36} \\
        \bottomrule
    \end{tabular*}
    \endgroup
\end{minipage}
\par\medskip

\par\medskip
\noindent\begin{minipage}{\linewidth}
    \centering
    \captionsetup{skip=5pt}
    \captionof{table}{\textbf{DDPM++ results under $v$-loss without augmentation.}}
    \label{tab:app-songunet-noaug-vloss}
    \begingroup
    \small
    \setlength{\tabcolsep}{3.2pt}
    \renewcommand{\arraystretch}{1.12}
    \begin{tabular*}{\linewidth}{@{\extracolsep{\fill}}llcrrrrrrrrr@{}}
        \toprule
        & & \multirow{2}{*}{$\lambda^\star$} & \multicolumn{3}{c}{$x$-pred} & \multicolumn{3}{c}{$v$-pred} & \multicolumn{3}{c}{$\epsilon$-pred} \\
        \cmidrule(lr){4-6}\cmidrule(lr){7-9}\cmidrule(lr){10-12}
        \multicolumn{2}{l}{Method} & & AUC & ASR & TPR & AUC & ASR & TPR & AUC & ASR & TPR \\
        \midrule
        Naive & ($q=20$) & $0.25$ & \underline{93.75} & \underline{86.19} & \underline{34.54} & 93.77 & 86.25 & 34.28 & 93.57 & 86.02 & 34.56 \\
        SecMI & ($q=12$) & fixed & 69.75 & 63.99 & 5.55 & 74.93 & 68.28 & 5.57 & 73.66 & 67.23 & 5.26 \\
        PIA & ($q=2$) & $0.25$ & 50.23 & 50.37 & 1.01 & \underline{95.73} & \underline{89.26} & 52.48 & \underline{96.30} & \underline{90.05} & 54.84 \\
        PIAN & ($q=2$) & $0.5$ & 50.36 & 50.39 & 1.06 & 94.31 & 88.44 & \underline{56.68} & 93.81 & 88.13 & \underline{58.36} \\
        \midrule
        TSC & ($q=15$) & $0.5$ & \textbf{98.04} & \textbf{93.03} & \textbf{81.31} & \textbf{97.89} & \textbf{92.83} & \textbf{80.56} & \textbf{97.88} & \textbf{92.71} & \textbf{80.68} \\
        \bottomrule
    \end{tabular*}
    \endgroup
    \medskip
    \captionof{table}{\textbf{DDPM++ results under $v$-loss with augmentation.}}
    \label{tab:app-songunet-aug-vloss}
    \begingroup
    \small
    \setlength{\tabcolsep}{3.2pt}
    \renewcommand{\arraystretch}{1.12}
    \begin{tabular*}{\linewidth}{@{\extracolsep{\fill}}llcrrrrrrrrr@{}}
        \toprule
        & & \multirow{2}{*}{$\lambda^\star$} & \multicolumn{3}{c}{$x$-pred} & \multicolumn{3}{c}{$v$-pred} & \multicolumn{3}{c}{$\epsilon$-pred} \\
        \cmidrule(lr){4-6}\cmidrule(lr){7-9}\cmidrule(lr){10-12}
        \multicolumn{2}{l}{Method} & & AUC & ASR & TPR & AUC & ASR & TPR & AUC & ASR & TPR \\
        \midrule
        Naive & ($q=20$) & $0.25$ & \underline{69.69} & \underline{64.32} & \underline{5.97} & 69.77 & 64.35 & 5.88 & 70.04 & 64.59 & 5.96 \\
        SecMI & ($q=12$) & fixed & 50.53 & 50.54 & 1.13 & 50.64 & 50.70 & 1.09 & 51.04 & 50.88 & 1.16 \\
        PIA & ($q=2$) & $0.25$ & 50.26 & 50.38 & 1.00 & \underline{74.43} & \underline{68.35} & \underline{8.32} & \underline{74.51} & \underline{68.29} & 6.60 \\
        PIAN & ($q=2$) & $0.5$ & 50.25 & 50.38 & 1.13 & 69.94 & 65.59 & 8.18 & 69.19 & 65.24 & \underline{8.31} \\
        \midrule
        TSC & ($q=15$) & $0.5$ & \textbf{80.80} & \textbf{73.97} & \textbf{19.42} & \textbf{80.73} & \textbf{73.95} & \textbf{18.45} & \textbf{80.94} & \textbf{74.20} & \textbf{19.75} \\
        \bottomrule
    \end{tabular*}
    \endgroup
\end{minipage}
\par\medskip

\par\medskip
\noindent\begin{minipage}{\linewidth}
    \centering
    \captionsetup{skip=5pt}
    \captionof{table}{\textbf{DDPM++ results under $\epsilon$-loss without augmentation.}}
    \label{tab:app-songunet-noaug-eloss}
    \begingroup
    \small
    \setlength{\tabcolsep}{3.2pt}
    \renewcommand{\arraystretch}{1.12}
    \begin{tabular*}{\linewidth}{@{\extracolsep{\fill}}llcrrrrrrrrr@{}}
        \toprule
        & & \multirow{2}{*}{$\lambda^\star$} & \multicolumn{3}{c}{$x$-pred} & \multicolumn{3}{c}{$v$-pred} & \multicolumn{3}{c}{$\epsilon$-pred} \\
        \cmidrule(lr){4-6}\cmidrule(lr){7-9}\cmidrule(lr){10-12}
        \multicolumn{2}{l}{Method} & & AUC & ASR & TPR & AUC & ASR & TPR & AUC & ASR & TPR \\
        \midrule
        Naive & ($q=20$) & $1.5$ & \underline{94.67} & 88.42 & 27.52 & 95.16 & 89.15 & 29.78 & 94.93 & 88.77 & 28.29 \\
        SecMI & ($q=12$) & fixed & 94.64 & 88.24 & 29.25 & 96.24 & 90.68 & 35.40 & 95.42 & 89.43 & 30.82 \\
        PIA & ($q=2$) & $1.75$ & 63.63 & 66.81 & 31.69 & \underline{98.65} & \underline{94.68} & 74.61 & \underline{98.72} & \underline{94.77} & 75.85 \\
        PIAN & ($q=2$) & $1.75$ & 93.08 & \underline{90.53} & \underline{72.11} & 95.29 & 92.41 & \underline{77.77} & 94.60 & 91.88 & \underline{76.73} \\
        \midrule
        TSC & ($q=15$) & $1.75$ & \textbf{99.37} & \textbf{96.68} & \textbf{91.22} & \textbf{99.52} & \textbf{97.18} & \textbf{93.22} & \textbf{99.49} & \textbf{97.06} & \textbf{92.75} \\
        \bottomrule
    \end{tabular*}
    \endgroup
    \medskip
    \captionof{table}{\textbf{DDPM++ results under $\epsilon$-loss with augmentation.}}
    \label{tab:app-songunet-aug-eloss}
    \begingroup
    \small
    \setlength{\tabcolsep}{3.2pt}
    \renewcommand{\arraystretch}{1.12}
    \begin{tabular*}{\linewidth}{@{\extracolsep{\fill}}llcrrrrrrrrr@{}}
        \toprule
        & & \multirow{2}{*}{$\lambda^\star$} & \multicolumn{3}{c}{$x$-pred} & \multicolumn{3}{c}{$v$-pred} & \multicolumn{3}{c}{$\epsilon$-pred} \\
        \cmidrule(lr){4-6}\cmidrule(lr){7-9}\cmidrule(lr){10-12}
        \multicolumn{2}{l}{Method} & & AUC & ASR & TPR & AUC & ASR & TPR & AUC & ASR & TPR \\
        \midrule
        Naive & ($q=20$) & $1.5$ & 71.95 & 66.28 & 6.39 & 72.58 & 66.68 & 6.36 & 72.99 & 67.00 & 6.91 \\
        SecMI & ($q=12$) & fixed & 64.86 & 60.36 & 4.16 & 67.70 & 62.54 & 4.69 & 67.46 & 62.39 & 4.79 \\
        PIA & ($q=2$) & $1.75$ & 53.51 & 53.11 & 3.44 & \underline{80.70} & \underline{73.49} & 13.71 & \underline{81.23} & \underline{74.05} & 13.36 \\
        PIAN & ($q=2$) & $1.75$ & \underline{72.38} & \underline{70.19} & \underline{13.50} & 75.17 & 71.33 & \underline{14.36} & 74.14 & 70.92 & \underline{15.54} \\
        \midrule
        TSC & ($q=15$) & $1.75$ & \textbf{88.53} & \textbf{81.54} & \textbf{22.41} & \textbf{89.20} & \textbf{82.26} & \textbf{24.12} & \textbf{89.35} & \textbf{82.48} & \textbf{23.28} \\
        \bottomrule
    \end{tabular*}
    \endgroup
\end{minipage}
\par\medskip

\subsection{Reference Calibration Across Noise Levels}
\label{app:rmc-noise-sweep}

We evaluate RMC across 45 noise levels for all three realistic Stable Diffusion
fine-tunes in Table~\ref{tab:reference-calibration-real}, using the same target
checkpoints. The nominal log-SNR grid spans
$[-5.5,5.5]$ in increments of $0.25$, with each coordinate mapped to the nearest
available diffusion timestep. At each level, RMC uses one shared noise draw
for the fine-tuned and pretrained models ($q=2$), without captions. We also report the uncalibrated fine-tuned loss from the same draw
($q=1$) to show the effect of subtracting the reference loss. 

\makeatletter
\setlength{\@fptop}{0pt}
\makeatother
\begin{figure}[!t]
    \centering
    \includegraphics[width=\linewidth]{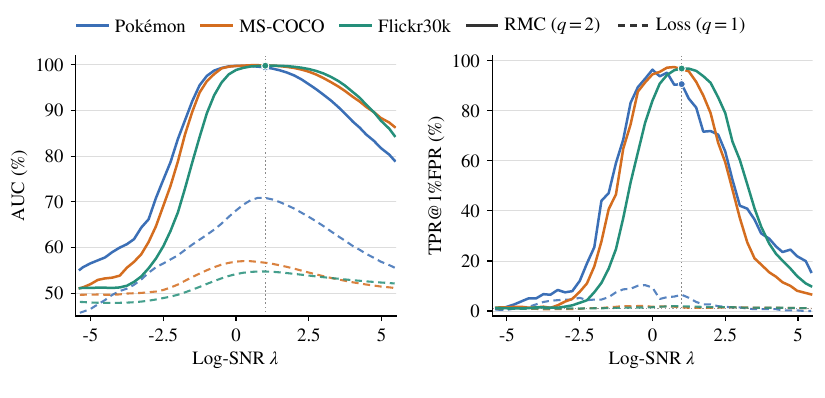}
    \caption{\textbf{Reference calibration across noise levels.}
    AUC (left) and TPR@1\%FPR (right) for the realistic Pok\'emon, MS-COCO, and
    Flickr30k fine-tunes. Solid curves show RMC ($q=2$). Dashed curves show
    the uncalibrated fine-tuned loss from the same noise draw ($q=1$).
    Each point uses a fixed score direction, with lower loss or calibrated
    residual indicating membership. The vertical guide marks $\lambda=1$,
    the operating point reported in Table~\ref{tab:reference-calibration-real}.}
    \label{fig:rmc-noise-sweep}
\end{figure}

\Needspace{4\baselineskip}
Around the operating region reported in Table~\ref{tab:reference-calibration-real},
RMC substantially improves both metrics over the uncalibrated loss on all three
datasets. The curves also characterize the sensitivity to the chosen noise level:
reference calibration reduces difficulty-related variation, while the available
membership signal can still vary along the diffusion trajectory.